\documentclass[journal,hidelinks]{IEEEtran}
\usepackage{amsmath,amsfonts}

\usepackage{array}
\usepackage[caption=false,font=normalsize,labelfont=sf,textfont=sf]{subfig}
\usepackage{textcomp}
\usepackage{stfloats}
\usepackage{url}
\usepackage{verbatim}
\usepackage{graphicx}
\usepackage{color}
\usepackage{bm}
\usepackage{amsmath}
\usepackage{amssymb}
\usepackage{mathtools}
\usepackage{amsthm}
\usepackage{amsfonts,bm}

\newcommand{\defeq}{:=}

\renewcommand{\xi}{\mX[i,:]}

\def\eqref#1{equation~\ref{#1}}

\def\1{\bm{1}}

\def\va{{\bm{a}}}
\def\vb{{\bm{b}}}

\def\vt{{\bm{t}}}

\def\vx{{\bm{x}}}
\def\vy{{\bm{y}}}

\def\mE{{\bm{E}}}

\def\mM{{\bm{M}}}

\def\mP{{\bm{P}}}

\def\mU{{\bm{U}}}

\def\mX{{\bm{X}}}
\def\mY{{\bm{Y}}}

\DeclareMathAlphabet{\mathsfit}{\encodingdefault}{\sfdefault}{m}{sl}
\SetMathAlphabet{\mathsfit}{bold}{\encodingdefault}{\sfdefault}{bx}{n}

\usepackage[numbers]{natbib}
\usepackage{algorithm}
\usepackage{algpseudocode}
\usepackage{float}
\usepackage{multirow}
\usepackage{multicol}
\usepackage{hyperref}       
\usepackage{booktabs}       
\usepackage{chngcntr}
\usepackage{makecell}

\begin{document}

\title{Towards Generalising Neural Topical Representations}


\author{Xiaohao Yang, He Zhao, Dinh Phung and Lan Du
\thanks{Xiaohao Yang, Dinh Phung and Lan Du are with Department of Data Science and AI, Faculty of IT, Monash University. E-mail: \{xiaohao.yang, dinh.phung, lan.du\}@monash.edu}
\thanks{He Zhao is with CSIRO’s Data61. E-mail: he.zhao@data61.csiro.au}}

\markboth{Journal of \LaTeX\ Class Files,~Vol.~14, No.~8, August~2021}%
{Shell \MakeLowercase{\textit{et al.}}: A Sample Article Using IEEEtran.cls for IEEE Journals}


\maketitle

\begin{abstract}
    Topic models have evolved from conventional Bayesian probabilistic models to recent Neural Topic Models (NTMs). Although NTMs have shown promising performance when trained and tested on a specific corpus, their generalisation ability across corpora has yet to be studied. In practice, we often expect that an NTM trained on a source corpus can still produce quality topical representation (i.e., latent distribution over topics) for the document from different target corpora to a certain degree. In this work, we aim to improve NTMs further so that their representation power for documents generalises reliably across corpora and tasks. To do so, we propose to enhance NTMs by narrowing the semantic distance between similar documents, with the underlying assumption that documents from different corpora may share similar semantics. Specifically, we obtain a similar document for each training document by text data augmentation. Then, we optimise NTMs further by minimising the semantic distance between each pair, measured by the Topical Optimal Transport (TopicalOT) distance, which computes the optimal transport distance between their topical representations. Our framework can be readily applied to most NTMs as a plug-and-play module. Extensive experiments show that our framework significantly improves the generalisation ability regarding neural topical representation across corpora. Our code and datasets are available at: \href{https://github.com/Xiaohao-Yang/Topic_Model_Generalisation}{https://github.com/Xiaohao-Yang/Topic\_Model\_Generalisation}.
\end{abstract}


\section{Introduction}
\IEEEPARstart{T}{opic} modelling is a powerful technique for discovering semantic structures of text corpora in an unsupervised manner. It brings success to various applications, such as information retrieval \citep{blei2003modeling}, marketing analysis \citep{reisenbichler2019topic}, social media analysis \citep{laureate2023systematic}, bioinformatics \citep{liu2016overview} and etc. Conventional topic models such as Latent Dirichlet allocation (LDA) \citep{blei2003latent} are Bayesian probabilistic models that assume generative stories of the data. With the increasing scale of data and the development of modern deep learning techniques, the union of deep learning and topic modelling, namely the Neural Topic Model (NTM) \citep{zhao2020neural}, is becoming a popular technique for text analytics. 

Given a collection of documents, a topic model learns a set of latent topics, each describing an interpretable semantic concept. A topic model is usually used in two ways: Using the topics to interpret the content of a corpus and using the topic distribution of a document as the semantic representation (i.e., topical representation). For the latter, the learned topical representations by topic models have shown good performance in downstream applications such as document classification, clustering, and retrieval. In practice, it is important that a trained model yields good representations for new documents. Ideally, these new documents are i.i.d samples from the same distribution of the training documents (e.g., from the same corpus). However, this assumption is usually too strong for real-world applications, where new documents may not share the same data distribution with the training data (e.g. from different corpora). In this work, given an NTM trained on a source corpus, we are interested in how well its power of learning neural topical representation of documents generalises to an unseen corpus without retraining. More importantly, we aim to propose a model-agnostic training scheme that can improve the generalisation power of an arbitrary NTM. Although many methods have been proposed for generalising deep neural networks to unseen domains \citep{wang2022generalizing,zhou2022domain}, most of them are designed for image data and cannot be applied to topic models. This is potentially because that topic models are unsupervised methods whose latent representations (topics) encode specific semantic meanings and the evaluation of a model's generalisation power is quite different from that of computer vision.
Therefore, we believe that the problem studied in this work has not been carefully investigated in the literature.

Our idea is straightforward: If an NTM generalises, it shall yield similar topical representations for documents with similar content. Based on this assumption, we further enhance NTMs by minimising the distance between similar documents, which are created by text data augmentation \citep{wei2019eda,shorten2021text, feng2021survey,bayer2022survey} during the training. Specifically, a document can be encoded as a latent distribution $\bm{z}$ over topics (i.e., topic distribution/topical representation) by NTMs. To make the model capable of producing quality $\bm{z}$ for unseen documents, we encourage the model to learn similar $\bm{z}$ for similar documents, which can be generated by document augmentations \citep{shorten2021text,bayer2022survey} such as adding, dropping and replacing words or sentences in the documents.
To bring the topical representations of similar documents close together, we need to measure the distance between topical representations. This is done by a topical optimal transport distance that computes the distance between documents' topic distribution $\bm{z}$. It naturally incorporates semantic information from topics and words into the distance computation between documents. Finally, with the optimal transport distance between the document and its augmentation, we propose to minimise this distance as a regularisation term for training NTMs for better generalisation.
Our generalisation regularisation (Greg) term can be easily plugged into the training procedure of most NTMs. Our main contributions are summarised as followings:
\begin{itemize}
    \item We are the first study of improving NTMs' generalisation capability regarding document representation, which is expected in practice, especially for downstream tasks based on document representation.
    \item We introduce a universal regularisation term for NTMs by applying text data augmentation and optimal transport, which brings consistent improvements over the generalisation ability of most NTMs.
    \item We examine the generalisation capability of NTMs trained on a source corpus by testing their topical representations on a different target corpus, which is a new setup for topic model evaluation. 
\end{itemize}

\section{Background}\label{background}
This section introduces the background of neural topic models and optimal transport, along with the notations used in this paper, which will facilitate the understanding of our method discussed in section \ref{method}. 
A summary of common math notations used in this paper is provided in Table \ref{math_notations}.

\subsection{Neural Topic Models}
Given a document collection, a topic model aims to learn the latent topics and the topical representation of documents. Specifically, a document in a text corpus $\mathcal{D}$ can be represented as a Bag-Of-Words (BOW) vector $\bm{x} \in \mathbb{N}^V$, where $V$ denotes the vocabulary size (i.e., the number of unique words in the corpus). A topic model learns a set of $K$ topics $\bm{T}\defeq\{\bm{t}_1,...,\bm{t}_K\}$ of the corpus, each $\bm{t}_k \in \Delta^{V}$ is a distribution over the $V$ vocabulary words. The model also learns a distribution over the $K$ topics $\bm{z} \in \Delta^{K}$ for document $\bm{x}$, by modelling $p(\bm{z}|\bm{x})$. $\bm{z}$ can be viewed as the topical representation of document $\bm{x}$. To train a topic model, one usually needs to ``reconstruct'' the BOW vector by modelling $p(\bm{x}|\bm{z})$.

Most conventional topic models such as LDA \citep{blei2003latent} are Bayesian probabilistic models, where $p(\bm{x}|\bm{z})$ is built with probabilistic 
graphical models and inferred by a dedicated inference process.
Alternatively, Neural Topic Models (NTMs)~\citep{zhao2021topic} have been recently proposed, which use deep neural networks to model $p(\bm{z}|\bm{x})$ and $p(\bm{x}|\bm{z})$.
Although there have been various frameworks for NTMs, models based on Variational Auto-Encoders (VAEs) \citep{kingma2013auto} and Amortised Variational Inference (AVI) \citep{rezende2014stochastic} are the most popular ones.

For VAE-NTMs, $p_{\phi}(\bm{x}|\bm{z})$ is modelled by a decoder network $\phi$: $\bm{x'} = \phi(\bm{z})$; $p(\bm{z}|\bm{x})$ is approximated by the variational distribution $q_\theta(\bm{z}|\bm{x})$ which is modelled by an encoder network $\theta$: $\bm{z} = \theta(\bm{x})$.
The learning objective of VAE-NTMs is maximising the Evidence Lower Bound (ELBO):
\begin{equation} \label{eq1}
\max_{\theta,\phi}{(\mathbb{E}_{q_{\theta}(\bm{z}|\bm{x})}[\log{p_{\phi}(\bm{x}|\bm{z})}]}-\mathbb{KL}[q_{\theta}(\bm{z}|\bm{x})\parallel p(\bm{z})]),
\end{equation}
where the first term is the conditional log-likelihood, and the second is the Kullback–Leibler (KL) divergence between the variational distribution of $\bm{z}$ and its prior distribution $p(\bm{z})$.
By using one linear layer for the decoder in NTMs, one can obtain the topic over word distributions by normalising the columns of the decoder's weight $W \in \mathbb{R}^{V\times K}$.
Note that although VAE-NTMs are of the most interest in this paper, our proposed method is not specifically designed for them, it can be applied to other NTM frameworks as well.

\begin{table}[!t]
  \caption{Summary of math notations}
  \label{math_notations}
  \centering
  \begin{tabular}{p{7mm}c|p{4.5cm}}
    \toprule
    Category & Notation & Description \\
    \midrule
    \multirow{7}{*}{NTM} 
    & $K, V, I, B$              & Number of topics, vocabulary words, topic top words, batch size;\\
    \rule{0pt}{2ex}
    & $\bm{d},\bm{x}, \bm{x}^s,\bm{x}^{aug}$   & Document, BOW vector, BOW  of source document and its augmentation;\\
    \rule{0pt}{2ex}
    & $\bm{z}, \bm{z}^s,\bm{z}^{aug}$ & Topical representation, topical representation of source document and its augmentation;\\
    \rule{0pt}{2ex}
    & $\bm{X}, \bm{X}^s, \bm{X}^{aug}$ & Batch of BOW vectors, Batch of BOW of source documents and their augmentations;\\
    \rule{0pt}{2ex}
    & $\bm{Z}, \bm{Z}^s, \bm{Z}^{aug}$ & Batch of topical representations, Batch of topical representations of source documents and their augmentations;\\
    \rule{0pt}{2ex}
    & $\mathcal{D}, \mathcal{D}^S, \mathcal{D}^T$ & Text corpus, the source and target corpus;\\
    \rule{0pt}{2ex}
    & $\mathcal{V}, \mathcal{V}^S, \mathcal{V}^T$ & Vocabulary, source and target vocabulary;\\
    \rule{0pt}{2ex}
    & $\bm{T}, \bm{\Tilde{T}}, \bm{t}_k,\bm{\Tilde{t}}_k$ & Set of topics, set of estimated topics, topic $k$, and estimated topic $k$;\\
    \rule{0pt}{2ex}
    & $\theta, \phi, W$ & Encoder network,  decoder network, and decoder weight;\\
    \midrule
    \multirow{7}{*}{OT} 
    & $\mX,\mY$              & Supports of two discrete distributions;\\
    \rule{0pt}{2ex}
    & $\va,\vb$   & Probability vectors;\\
    \rule{0pt}{2ex}
    & $\Delta^{M}$ & $M$-dimensional probability simplex;\\
    \rule{0pt}{2ex}
    & $D,D_{\mM},D_{\mM,\lambda}$ & General distance, OT distance, Sinkhorn distance;\\
    \rule{0pt}{2ex}
    & $\mM,\bm{M^t},\bm{M^d}$ & General cost matrix, topic cost matrix, document cost matrix;\\
    \rule{0pt}{2ex}
    & $\mP,\mU$ & Transport matrix, transport polytope;\\
    \midrule
    \multirow{3}{*}{General} 
    & $\bm{E}, \bm{e}, L$              & Word embedding matrix, word embedding vector, embedding dimension;\\
    \rule{0pt}{2ex}
    & $\beta,\gamma,\lambda$ & Augmentation strength, regularisation weight, Sinkhorn hyperparameter;\\
    \rule{0pt}{2ex}
    & $\mathcal{F}, f_N,f_I$   & Data augmentation function, normalising function, function return top $I$ elements\\
  \bottomrule
\end{tabular}
\end{table}

\subsection{Optimal Transport}
Optimal Transport (OT) has been widely used in machine learning for its capability of comparing probability distributions \citep{peyre2019computational}. Here, we focus on the case of OT between discrete distributions. 
Let  $\mu(\mX, \va) \defeq \sum_{i=1}^N \va_i \delta_{\vx_i}$ and $\mu(\mY, \vb) \defeq \sum_{j=1}^{M} \vb_j \delta_{\vy_j}$, where $\mX =\{\vx_1, \cdots, \vx_N\}$ and $\mY = \{\vy_1, \cdots, \vy_M\}$ denote the supports of the two distributions, respectively; 
$\va \in \Delta^{N}$ and $\vb \in \Delta^{M}$ are probability vectors in $\Delta^N$ and $\Delta^M$, respectively. The OT distance between $\mu(\mX,\bm{a})$ and $\mu(\mY,\bm{b})$ can be defined as:
\begin{equation}
\label{eq-def-ot}
D_{\mM}\left(\mu(\mX,\bm{a}), \mu(\mY,\bm{b})\right) 
\defeq
\inf_{\mP \in U(\va, \vb)} \langle \mP , \mM \rangle,
\end{equation}
where $\langle\cdot,\cdot\rangle$ denotes the Frobenius dot-product;
$\mM \in \mathbb{R}_{\ge 0}^{N \times M}$ is the cost matrix of the transport which defines the
pairwise cost between the supports;
$\mP \in \mathbb{R}_{>0}^{N \times M}$ is the transport matrix;
$U(\va, \vb)$ denotes the transport polytope of $\va$ and $\vb$,
which is the polyhedral set of $N \times M$ matrices:
\begin{equation}
    \mU(\va, \vb) \defeq \left\{\mP \in \mathbb{R}^{N \times M}_{>0} | \mP \boldsymbol{1}_{M} = \va, \mP^T \boldsymbol{1}_{N} = \vb\right\},
\end{equation}
where $\boldsymbol{1}_{M}$ and $\boldsymbol{1}_{N}$ are the $M$ and $N$-dimensional column vector of ones, respectively.

The OT distance can be calculated by finding the optimal transport plan $\bm{P}^*$, and various OT solvers have been proposed \citep{flamary2021pot}. The direct optimisation of Eq.\ (\ref{eq-def-ot}) is computationally expensive, \cite{cuturi2013sinkhorn} introduced the entropy-regularised OT distance, known as the Sinkhorn distance, which is more efficient for large-scale problems. It can be defined as:
\begin{equation} \label{sinkhorn}
D_{\bm{M},\lambda}(\mu(\mX,\bm{a}), \mu(\mY,\bm{b}))
\defeq
\inf_{\bm{P}\in U_{\lambda}(\bm{a},\bm{b})}\langle \mP , \mM \rangle,
\end{equation}
where $\mU_{\lambda}(\bm{a},\bm{b})$ defines the transport plan with the constraint that: $h(\bm{P})\geq h(\bm{a})+h(\bm{b})-\lambda$, where $h(\cdot)$ denotes the entropy function and $\lambda \in [0,\infty]$ is the hyperparameter.

\section{Method}\label{method}
\subsection{Problem Setting}
In this paper, given an NTM trained on a source corpus $\mathcal{D}^S$, we are interested in how to train a neural topic model on $\mathcal{D}^S$ so that it can generate good topical representations not only on $\mathcal{D}^S$ but also on unseen corpora \textit{without} retraining on them.
%
%
We note that an NTM with a standard
training scheme usually has some intrinsic generalisation power to new corpora, as documents from
different domains may share similar semantics. However, we argue that such intrinsic generalisation power is not enough to generate quality topical representations for unseen documents.
In this work, we aim to study and improve such intrinsic power of NTMs.

\begin{figure*}[!ht]
    \centering
\includegraphics[width=0.8\textwidth,height=\textheight,keepaspectratio]{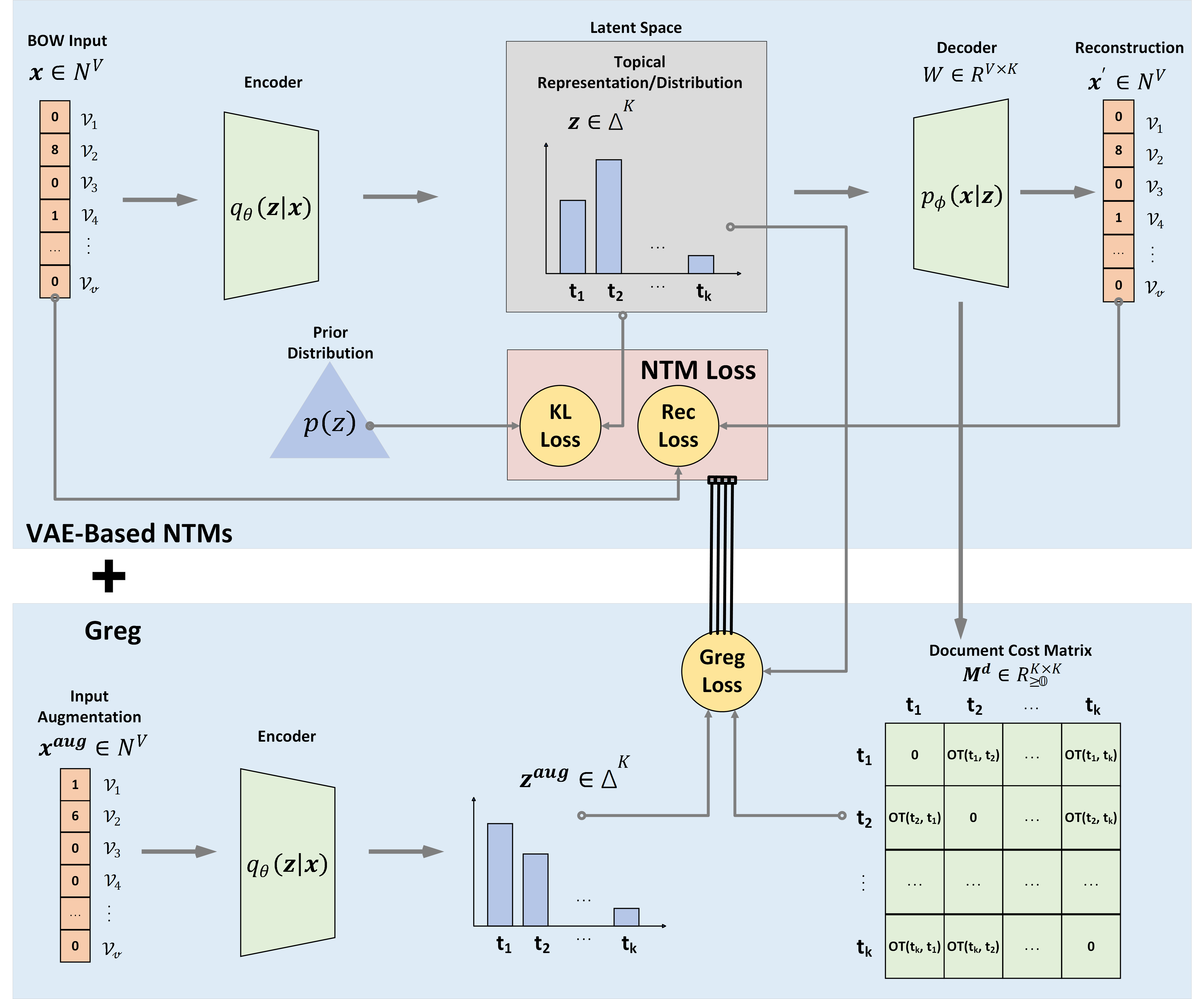}
    \caption{Neural Topic Model (NTM) with Generalisation Regularisation (Greg). The BOW vectors of a document and its augmentation are encoded as the topical representations, respectively; Besides the common VAE-NTMs that aim to reconstruct the input (``Rec Loss'') and match the posterior distribution to the prior (``KL Loss''), we encourage the model to produce a similar $\bm{z}$ for the original document and its augmentation; The distance between $\bm{z}$ is measured by TopicalOT as our ``Greg Loss'', which is guided by the document cost matrix whose entries specify the OT cost of moving between topics. Our framework can be readily applied to most NTMs as a plug-and-play module. Note: We draw two encoders here for tidy illustration; they are identical.}
    \label{fig_framework}
\end{figure*}

\subsection{Overview of the Proposed Method}
We enhance NTMs' generalisation by assuming that a document’s topical representation should be close to the topical representations of its augmentations.
Specifically, let $\bm{x}^s$ be the BOW vector of the document from the source corpus. Suppose a stochastic function $\mathcal{F}$ can produce a random augmentation $\bm{x}^{aug}$ that is semantically similar to $\bm{x}^s$: $\bm{x}^{aug} \defeq \mathcal{F}(\bm{x}^s)$. As both $\bm{x}^s$ and $\bm{x}^{aug}$ share similar semantics, their topical representations $\bm{z}^s$ and $\bm{z}^{aug}$ should stay close in the learned representation space. 
To achieve this, we introduce a new semantics-driven regularisation term to the existing training objective of an NTM, which additionally minimises the distance between $\bm{z}^s$ and $\bm{z}^{aug}$ 
with respect to the encoder parameters $\theta$:
\begin{equation} \label{general_form}
\min_{\theta}{D(\bm{z}^s, \bm{z}^{aug})}.
\end{equation}


Straightforward choices of $D$ can be the Euclidean, Cosine, or Hellinger distances, to name a few.
However, these distances cannot sufficiently capture the semantic distance between documents in
terms of their topic distributions. To address this issue, we propose to use a topical optimal transport distance
between documents, inspired by the Hierarchical OT (HOT) distance from \cite{yurochkin2019hierarchical}. We refer to it as TopicalOT throughout our paper for clarity within our context. Specifically, we are given the word embedding matrix $\bm{E} \in \mathbb{R}^{V\times L}$ from pre-trained models such as Word2Vec \citep{mikolov2013efficient}, GloVe \citep{pennington2014glove}, BERT \citep{devlin2018bert} and etc., where $V$ is the vocabulary size and $L$ is the embedding dimension.
Each word embedding is denoted by $\bm{e}^{v}\in\mathbb{R}^L$, where $v \in [1,\dots, V]$ denotes the $v$-th word in the vocabulary.
Let $\bm{T}\defeq\{\bm{t}_1,...,\bm{t}_K\}\subset{\Delta}^V$ be the learned topics of the corpus, each of which is a distribution over words. 
Therefore, each topic $\vt_k$ can be viewed as a discrete distribution whose supports are the word embedding: 
$\mu(\mE, \vt_k)$.
Then, the OT distance between topic $\bm{t}_{k_1}$ and $\bm{t}_{k_2}$ can be computed by:
\begin{equation} \label{topic_distance}
\begin{aligned}
& \text{WMD}(\vt_{k_1},\vt_{k_2}) \defeq
D_{\bm{M^t}}(\mu(\bm{E},\vt_{k_1}),\mu(\bm{E},\vt_{k_2})), \\
& \text{where }\bm{M^t}_{v_1,v_2} \defeq 1-\text{cos}{(\bm{e}^{v_1},\bm{e}^{v_2})},
\end{aligned}
\end{equation} 
where $D_{\bm{M^t}}(\cdot,\cdot)$ denotes the OT distance in Eq.\ (\ref{eq-def-ot}); $\bm{M^t}\in\mathbb{R}_{\ge 0}^{V\times V}$ is the topic cost matrix and $\text{cos}(\cdot,\cdot)$ denotes the Cosine similarity; $k_1,k_2 \in [1,...,K]$ are indices of topics and  $v_1,v_2 \in [1,...,V]$ are indices of vocabulary words. Eq.\ (\ref{topic_distance}) can be regarded as the Word Mover's Distance (WMD) \citep{kusner2015word}, measuring the topic distance instead of the document distance. 

Similarly, as the document's topical representation $\bm{z}$ can be viewed as a discrete distribution whose supports are the topics, the OT Distance between documents $\bm{d}_i$ and $\bm{d}_j$ is computed by:
\begin{equation} \label{HOTT}
\begin{aligned}
    & \text{TopicalOT}(\bm{d}_i,\bm{d}_j) \defeq
    D_{\bm{M^d}}(\mu(\bm{T},\bm{z}_{i}),\mu(\bm{T},\bm{z}_j)) \\
    & \text{where }\bm{M^d}_{k_1,k_2} \defeq \text{WMD}(\vt_{k_1},\vt_{k_2}),
\end{aligned}
\end{equation}
where $\bm{M^d}\in\mathbb{R}_{\ge 0}^{K\times K}$ is the document cost matrix whose entries indicate the OT distance between topics as in Eq.\ (\ref{topic_distance}). 


As for topics, we use the decoder weights: $W\in \mathbb{R}^{V\times K}$ as the representation of topics, like other NTMs. Since OT measures the distance between distributions, we normalise each topic as the topic over word distribution by the softmax function: 
\begin{equation} \label{eq_topic}
\bm{t}_k\defeq\text{softmax}(W^{\text{T}}_{k,:}), 
\end{equation}
where $\text{T}$ denotes the matrix transpose operation. Then, we can construct $\bm{M^d}$ by computing the OT distance between topics as Eq.\ (\ref{topic_distance}). Putting all together, we have:
\begin{subequations}\label{full_form}
\begin{align}
    & \min_{\theta, W}{D_{\bm{M^d}}}(\mu(\bm{T},\bm{z}^s),\mu(\bm{T},\bm{z}^{aug})), \label{full_form_1} \\
    & \text{where } \bm{T}\defeq \{\bm{t}_1,...,\bm{t}_K\},\ \bm{z}\defeq\theta(\bm{x}), \nonumber \\
    & \bm{M^d}_{k_1,k_2} \defeq D_{\bm{M^t}}(\mu(\bm{E},\bm{t}_{k_1}),\mu(\bm{E},\bm{t}_{k_2})), \label{full_form_2} \\
    & \bm{M^t}_{v_1,v_2} \defeq 1-\cos{(\bm{e}^{v_1},\bm{e}^{v_2})}. \label{full_form_3}
\end{align}
\end{subequations}

Intuitively, we are encouraging the model to produce similar topical representations for documents with similar semantics, where the semantic similarity between topical representations is captured by a two-level OT distance. At the document level, TopicalOT compute the OT distance between two topical representations of documents, where the cost matrix is defined by the semantic distances between topics. At the topic level, the distance between two topics is again measured by OT, where the transport cost is determined by distances between word embeddings. The whole computation injects rich semantic information from both external (i.e., word embeddings from the pre-trained model) and internal (i.e., the topic model itself) sources, thus better capturing the semantic similarity between documents.

\subsection{Efficiently Computing Topical OT}
Directly solving the problem in Eq.\ (\ref{full_form}) during training is computationally expensive for two reasons: (i) To obtain $\bm{M^d}$ in Eq.\ (\ref{full_form_2}), we need to compute the OT distance between $K\times (K-1)/2$ topic pairs, each of them has a problem space of $V\times V$. This will be expensive since a corpus may have a large vocabulary size; (ii) NTMs are usually trained on a large text corpus in batches of documents $\bm{X}\defeq\{\bm{x}_i\}^{B}_{i=1}\in\mathbb{N}^{B\times V}$ where $B$ denotes the batch size, so we need to compute between $\bm{Z}^s$ and $\bm{Z}^{aug}$ where $\bm{Z}\defeq\{\bm{z}_i\}^{B}_{i=1}\in\mathbb{R}^{B\times K}$. While the original HOT leverages the algorithm in \cite{bonneel2011displacement} for the computation of OT, which can not support the computation between $\bm{z}$ pairs of $\bm{Z}^s$ and $\bm{Z}^{aug}$ in parallel, thus causing an enormous computational cost during training. 

To address the first issue, we leverage the fact that a topic's semantics can be captured by its most important words. Specifically, when computing $\bm{M^d}$, we reduce the dimension of each topic $\bm{t}_k$ by considering only the top $I$ words that have the largest weights in the topic: 
\begin{equation} \label{eq_appro_topic}
\bm{\Tilde{t}}_k\defeq f_N(f_I((\bm{t}_{k})), 
\end{equation}
where $\bm{t}_k$ is the topic before approximation defined in Eq.\ (\ref{eq_topic}); $f_I$ is a function that returns the subset that contains $I$ elements with the largest weights; $f_N$ denotes the function for re-normalising by dividing by the sum. Now, we reduce the problem space of solving OT between one topic pair in Eq.\ (\ref{full_form_2}) from $V\times V$ to $I\times I$, and each estimated topic $\bm{\Tilde{t}}_k$'s related vocabulary and word embeddings become $\mathcal{V}^{\bm{\Tilde{t}}_k}$ and $\bm{E}^{\bm{\Tilde{t}}_k}$, respectively. Since the vocabulary of each topic become different then, the topic cost matrix in Eq.\ (\ref{full_form_3}) will vary for different topic pairs, which is denoted by $\bm{M}^{\bm{\Tilde{t}}_{k_1},\bm{\Tilde{t}}_{k_2}}\in \mathbb{R}^{I\times I}_{\geq 0}$ for topic $\Tilde{\bm{t}}_{k_1}$ and $\Tilde{\bm{t}}_{k_2}$. Now, we rewrite Eq.\ (\ref{full_form_2}) and Eq.\ (\ref{full_form_3}) as:
\begin{subequations}
\begin{align}
    & \bm{M^d}_{k_1,k_2} \defeq D_{\bm{M}^{\bm{\Tilde{t}}_{k_1},\bm{\Tilde{t}}_{k_2}}}(\mu(\bm{E}^{\bm{\Tilde{t}}_{k_1}},\bm{\Tilde{t}}_{k_1}),\mu(\bm{E}^{\bm{\Tilde{t}}_{k_2}},\bm{\Tilde{t}}_{k_2})), \\
    & \text{where } \bm{M}^{\bm{\Tilde{t}}_{k_1},\bm{\Tilde{t}}_{k_2}}_{v_1,v_2} \defeq 1-\cos{(\bm{e}^{v_1},\bm{e}^{v_2})}.
\end{align}
\end{subequations}

So far, we reduce the size of topic cost matrix to $I \times I$ (i.e., $\ v_1,v_2\in [1,...,I]$). We approximate this way because only a small subset of most important words is helpful for the understanding
of a topic, similar to the consideration when evaluating topic coherence \citep{newman2010automatic,lau2014machine}. 

To address the second issue, we replace the OT distance between $\bm{z}$ with the Sinkhorn distance defined in Eq.\ (\ref{sinkhorn}), and leverage the Sinkhorn algorithm \citep{cuturi2013sinkhorn} for its efficiency and parallelisation. As for the distance between topics, we keep it as OT distance because each topic does not share the same vocabulary set by our approximation approach, which results in different cost matrices for each topic pair. Although it has to be computed pairwisely for topic paris, it is still feasible since the number
of topics $K$ is usually small in NTMs. Putting all together, we have the following as the final form for efficiently
computing TopicalOT during training:
\begin{subequations}\label{appro_form}
\begin{align}
    & \min_{\theta, W}{D_{\bm{M^d},\lambda}}(\mu(\bm{\Tilde{T}},\bm{Z}^s),\mu(\bm{\Tilde{T}},\bm{Z}^{aug})), \label{appro_form_1} \\
    & \text{where }\bm{\Tilde{T}}\defeq  \{\bm{\Tilde{t}}_1,...,\bm{\Tilde{t}}_K\},\ \bm{Z}\defeq\theta(\bm{X}), \nonumber \\
    & \bm{M^d}_{k_1,k_2}\defeq D_{\bm{M}^{\bm{\Tilde{t}}_{k_1},\bm{\Tilde{t}}_{k_2}}}(\mu(\bm{E}^{\bm{\Tilde{t}}_{k_1}},\bm{\Tilde{t}}_{k_1}),\mu(\bm{E}^{\bm{\Tilde{t}}_{k_2}},\bm{\Tilde{t}}_{k_2})), \label{appro_form_2} \\
    & \bm{M}^{\bm{\Tilde{t}}_{k_1},\bm{\Tilde{t}}_{k_2}}_{v_1,v_2}\defeq 1-\cos{(\bm{e}^{v_1},\bm{e}^{v_2})}. \label{appro_form_3}
\end{align}
\end{subequations}
Compared to the original HOT distance, which only serves as a distance measure between documents, our approximation by TopicalOT supports the computation of distances between batches of document pairs in parallel. Additionally, it is optimisable, allowing for easy integration during model training.

\begin{table}[!t]
  \caption{Word-Level DAs}
  \label{DA_Summary}
  \centering
  \begin{tabular}{c|p{4.8cm}}
    \toprule
    DA & Description \\
    \midrule
    Random Drop              & Randomly sample $n$ words from the document and drop; \\
    \rule{0pt}{2ex}
    Random Insertion         & Randomly sample $n$ words from the vocabulary and add to the document; \\
    \rule{0pt}{2ex}
    Random to Similar        & Randomly replace $n$ words of the document with one of their top similar words within the vocabulary; \\
    \rule{0pt}{2ex}
    Highest to Similar       & Replace $n$ words with the highest Term Frequency–Inverse Document Frequency (TF-IDF) weight with one of their top similar words within the vocabulary; \\
    \rule{0pt}{2ex}
    Lowest to Similar        & Replace $n$ words with the lowest TF-IDF weight with one of their top similar words within the vocabulary; \\
    \rule{0pt}{2ex}
    Random to Dissimilar     & Randomly replace $n$ words of the document with one of their top dissimilar words within the vocabulary; \\
    \rule{0pt}{2ex}
    Highest to Dissimilar    & Replace $n$ words with the highest TF-IDF weight with one of their top dissimilar words within the vocabulary; \\
    \rule{0pt}{2ex}
    Lowest to Dissimilar     & Replace $n$ words with the lowest TF-IDF weight with one of their top dissimilar words within the vocabulary; \\
  \bottomrule
\end{tabular}
\end{table}

\subsection{Document Augmentation (DA)} As for the function $\mathcal{F}(\cdot)$ that generates a random augmentation $\bm{x}^{aug}$ of the original document $\bm{x}^s$, our framework is agnostic to the text augmentation approach employed and is not limited to those discussed in \cite{wei2019eda, shorten2021text, feng2021survey, bayer2022survey}. As summarised in \cite{bayer2022survey}, document augmentation can occur at the character, word, phrase, or document level. Since common NTMs are trained on BOWs, we focus on word-level augmentation, which can be efficiently integrated during training. Different word-level \citep{ma2019nlpaug} document augmentations are investigated and their descriptions are summarised in Table \ref{DA_Summary}. Their effect to our generalisation framework are studied in section \ref{DA_study}. As a general form, here we write $\bm{z}^{aug}$ is obtained by: 
\begin{equation}\label{DA}
\bm{z}^{aug}\defeq \text{softmax}(\theta(\mathcal{F}(\bm{x}^s,\beta,\Omega))),
\end{equation}
where $\beta$ is the augmentation strength that determines the number of words to be varied: $n = \text{ceil}(\beta \times l)$ where $l$ is the document length and $\text{ceil}(\cdot)$ rounds up the given number; $\Omega$ denotes other information needed for the augmentation, such as the number of top words for replacement, and the pre-trained word embeddings $\bm{E}$ to provide similarity between words.

\renewcommand{\algorithmicrequire}{Input:}
\renewcommand{\algorithmicensure}{Output:}
\begin{algorithm}[!t]
\caption{Neural Topic Model with Greg}\label{alg:1}
\begin{algorithmic}[1]
\Require Dataset $\mathcal{D}^s=\{\bm{x}_{i}\}_{i=1}^{N}$, Pre-trained word embeddings $\bm{E}$, Topic number K, Regularisation weight $\gamma$, Augmentation strength $\beta$
\Ensure $\theta,\phi$
\State Randomly initialise $\theta$ and $\phi$;
\While{Not converged}
    \State Sample a batch of data $\bm{X}$;
    \State Compute $\bm{Z}=\text{softmax}(\theta(\bm{X}))$;
    \State Compute $\bm{Z}^{aug}$ by Eq.\ (\ref{DA});
    \State Get topics $\bm{\Tilde{T}}$ from $\phi$ by Eq.\ (\ref{eq_topic}) and (\ref{eq_appro_topic});
    \For{each topic paris $\bm{\Tilde{t}}_{k_1}, \bm{\Tilde{t}}_{k_2}$}
    \State Construct $\bm{M}^{\bm{\Tilde{t}}_{k_1},\bm{\Tilde{t}}_{k_2}}$ in Eq.\ (\ref{appro_form_3});
    \State Compute $\bm{M^d}_{k_1,k_2}$ in Eq.\ (\ref{appro_form_2});
    \EndFor
    \State Compute the loss defined in Eq.\ (\ref{joint_training});
    \State Compute gradients w.r.t $\theta$ and $\phi$;
    \State Update $\theta$ and $\phi$ based on the gradients;
\EndWhile
\end{algorithmic}
\end{algorithm}

\subsection{Integrating Greg to the Training of Existing NTMs}
The integration of Greg with existing NTMs is illustrated in Figure \ref{fig_framework}. With the primary goal of NTMs that aims to reconstruct the input documents and match the posterior to the prior distribution, we propose the following joint loss:
\begin{equation}\label{joint_training}
\begin{aligned}
\min_{\theta, \phi} (  \gamma \cdot \mathbb{E}_{q_{\theta}(\bm{z}^s|\bm{x})}[D_{\bm{M^d},\lambda}(\mu(\bm{\Tilde{T}},\bm{z}^s), \mu(\bm{\Tilde{T}},\bm{z}^{aug}))] \\ + \mathcal{L}^{\text{NTM}}  ),
\end{aligned}
\end{equation}
where $\mathcal{L}^{\text{NTM}}$ is the original training loss of an NTM, which can be the ELBO in Eq.\ (\ref{eq1}) or other losses; The first term is the proposed Greg loss, where $\bm{z}^s\defeq\text{softmax}(\theta(\bm{x}^s))$; $\bm{z}^{aug}$ is obtained by Eq.\ (\ref{DA}); $\bm{M^d}$ is parameterised by $\phi$\footnote{Precisely, $\bm{M^d}$ is parameterised
by $W$ which is the weight of the linear layer of $\phi$.} and can be obtained by solving Eq.\ (\ref{appro_form_2}) and Eq.\ (\ref{appro_form_3}); $\gamma$ is the hyperparameter that determines the strength of the regularisation; $\lambda$ is the hyperparameter for the Sinkhorn distance. The training algorithm of our generalisation regularisation (Greg) is summarised in Algorithm \ref{alg:1}. Notably, both the Sinkhorn distance and OT distance support auto differentiation in deep learning frameworks such as PyTorch \citep{patrini2020sinkhorn,bonneel2011displacement}, thus the loss in Eq.\ (\ref{appro_form}) is differentiable in terms of $\theta$ and $\phi$.

\section{Related Work}
\subsection{Neural Topic Models}
For a comprehensive review of NTMs, we refer the readers to \cite{zhao2021topic}. Here, we mainly focus on models based on VAE \citep{kingma2013auto} and AVI \citep{rezende2014stochastic}. Early works of NTMs focus on studying the prior distributions for the latent variables, such as Gaussian \citep{miao2017discovering} and various approximations of the Dirichlet prior \citep{srivastava2017autoencoding,zhang2018whai,burkhardt2019decoupling} for its difficulty in reparameterisation \citep{tian2020learning}. Recent NTMs mainly leverage external information such as complementary metadata in \citep{card2017neural} and contextual embeddings in \citep{dieng2020topic,bianchi2020pre,bianchi2020cross,xu2022hyperminer}. In this work, we are interested in generalising NTMs instead of proposing a new NTM. And we believe that our method is general to improve the generalisation of most NTMs not limited to VAE-NTMs.

\subsection{Topic Models and Optimal Transport}
Recently, a few works have built the connection between
topic modelling and OT, most focusing on developing new OT frameworks for topic modelling, such as non-neural topic model \citep{huynh2020otlda} and NTMs \citep{zhao2020neural,nan2019topic,zhang2023topic}.
Our method is not an OT framework for NTMs but a general regularisation term to improve the generalisation of NTMs, which is also compatible with NTMs based on the OT frameworks.

\subsection{Topic Model Generalisation}
Model generalisation is a popular topic in machine learning. However,
the generalisation of topic models, especially NTMs, has not been comprehensively studied. (i) Most existing works focus on generalising topic models to multiple domains with access to the full or partial data of the new domains for retraining/fine-tuning, such as the models on
continual lifelong learning \citep{chen2014topic,chen2015lifelong,blum2016generalized,chen2019affinity,gupta2020neural,qin2021lifelong,zhang2022lifelong,lei2023nmtf} and few-shot learning \citep{iwata2021few,duan2022bayesian,xu2024context}. While our approach needs no access to the data in the new domains nor retraining of the model. (ii) Some approaches focus on the generalisation of topics across different languages under the zero-shot or few-short setting \citep{bianchi2020cross,chang2021word,grootendorst2022bertopic}. While ours focuses on the generalisation of topical representation of unseen documents. (iii) Recent Large Language Model (LLM) based topic models \citep{wang2023prompting,pham2023topicgpt,chang2024enhanced} extract or refine topics by prompting, which inherits the generalisation power of LLMs. They primarily focus on topics and pay less attention on document representations. Moreover, their generalisation capability originates from numerous training data, supervised fine-tuning, etc., of LLMs, which is different from our setting as we only train the NTM on the source domain. Moreover, they require word sequences of documents as input, which is different from conventional NTMs that use BOWs. (iv) In terms of domain generalisation \citep{wang2022generalizing,zhou2022domain}, extensive works exist for computer vision tasks, including image classification \citep{yue2019domain,liu2021learning}, semantic segmentation \citep{gong2019dlow,li2021semantic} and action recognition \citep{li2017domain,li2019episodic}; as well as some natural language processing tasks, such as sentiment classification \citep{balaji2018metareg,wang2020unseen} and semantic parsing \citep{wang2020meta}. To the best of our knowledge, there is no universal module or regulariser known to be applicable to topic models for domain generalisation, probably due to topic models' unsupervised nature and the lack of comprehensive evaluation. Ours is the first specialised for NTMs' generalisation. (v) Regarding the learning strategy, our work is related to the contrastive learning framework \citep{chen2020simple}, but we focus on only the positive pairs and leverage semantic distance between documents. As for the work that most related to ours, the Contrastive Neural Topic Model (CLNTM)  proposed in \citep{nguyen2021contrastive}  uses the contrastive distance \citep{chen2020big} to regularse the topical representagion of documents. There are fundamental differences between CLNTM and ours. Firstly, theirs does not focus on
the generalisation of NTMs. Secondly, the distance between document representations is measured by Cosine distance in CLNTM, while we use TopicalOT, which incorporates semantic information from topics and words. Besides the difference with the existing literature discussed above, our approach is expected to be a general regularisation to improve other NTMs.

\begin{table}[!t]
  \caption{Statistics of the Datasets}
  \label{data_statisc}
  \centering
  \begin{tabular}{ccccc}
    \toprule
    Dataset & \# Docs & Voc Size & Avg. Length & \# Labels\\
    \midrule
    20News              & 18846 & 1997   & 87 & 20 \\
    R8                & 7674  & 5047   & 56 & 8  \\
    Webs              & 12337 & 4523   & 14 & 8  \\
    TMN               & 32597 & 12004  & 18 & 7  \\
    DBpedia           & 19993 & 9830   & 23 & 14 \\
  \bottomrule
\end{tabular}
\end{table}

\section{Experiments}
\subsection{Experimental Settings}
\subsubsection{Datasets}
We conduct our experiments on five widely-used datasets: 20 Newsgroup (\textbf{20News}) \citep{lang1995newsweeder}, \textbf{R8}\footnote{https://www.kaggle.com/datasets/weipengfei/ohr8r52}, Web Snippets (\textbf{Webs}) \citep{phan2008learning}, Tag My News (\textbf{TMN}) \citep{vitale2012classification} and \textbf{DBpedia} \citep{zhang2015character}.
We pre-process the documents as BOW vectors by the following steps: We clean the documents by removing special characters and stop words, followed by tokenization. Then we build the vocabulary by considering the words with document frequency greater than five and less than 80\% of the total documents. As we use the pre-trained word embeddings of GloVe \citep{pennington2014glove} pre-trained on Wikipedia\footnote{https://nlp.stanford.edu/projects/glove/}, we filter the vocabulary words by keeping only the words that appear in the vocabulary set of GloVe. Finally, we convert documents to BOW vectors based on the final vocabulary set. The statistics of the pre-processed datasets are summarised in Table \ref{data_statisc}. These datasets are further randomly split as training and testing sets by 8:2 for our experiments.

\subsubsection{Evaluation Protocol}\label{eval_metrics}
\paragraph{Topical Representation Quality} We focus on the evaluation\footnote{All experiments in this paper are conducted five times with different random seeds. Mean and std values (in percentage) of metrics are reported.} of the quality of documents' topical representations, which is done by downstream tasks where the topical representations are used as input features for document classification and clustering: (i) Document Classification: We use a trained model to infer the topic distributions of the training and testing documents as their topical representations. Then we train a random forest classifier using the training documents' topical representation and evaluate the Classification Accuracy (\textbf{CA}) on testing documents. The random forest classifier consists of ten decision trees with a maximum depth of 8, which has the same setting as the previous work in \cite{nguyen2021contrastive}. (ii) Document Clustering: We evaluate the clustering performance of test documents' topical representation based on the widely-used Purity and Normalised Mutual Information (NMI). Following \cite{nguyen2015improving}, we assign each test document to a cluster corresponding to its top-weighted topic while computing Purity and NMI (denoted by Top-Purity (\textbf{TP}) and Top-NMI (\textbf{TN}), respectively) based on the documents' cluster assignments and their true labels (see Section 16.3 in \cite{schutze2008introduction} for calculation details).

\paragraph{Topical Representation Generalisation} To evaluate the generalisation of topic representations, we train a model on the source corpus and test on a target corpus.
We explore two configurations of the target corpus: (i) from a different domain of the source corpus; (ii) from the same domain of the source corpus but with noise. When the targets are different text corpora, they may not share the same vocabulary sets, as different pre-processing may result in various subsets of the language's vocabulary for specific text corpus. This makes NTM generalisation harder as an NTM can not accept input documents with a vocabulary set different from the training vocabulary. To address this issue, we unite the vocabulary set of all corpora for BOW representation during the training, allowing NTMs to accept input documents from different corpora (with varying vocabulary sets). When the target is the noisy source corpus, the noisy versions are created by randomly sampled compositions of document augmentations described in Table \ref{DA_Summary}, where the augmentation strength is set as 0.75 (i.e. changing 75\% of words in the original document). Notably, for all source-to-target tasks, the models are only trained on the source corpus with no access to the documents of the target corpus. 


\paragraph{Topic Quality} Although we are focusing on evaluating the generalisation capability of NTMs regarding the topical representation of documents, we also report the topic coherence by computing the commonly-used Normalised Pointwise Mutual Information (NPMI) score, following the standard protocol of evaluating topic models.
To evaluate the quality of learned topics, we compute NPMI \citep{aletras2013evaluating,lau2014machine} based on the top 10 words for each topic. Following \cite{yang2015efficient} and \cite{zhao2018dirichlet}, we compute the average NPMI score of the top coherent topics with the highest NPMI since not all learned topics are interpretable. Notably, we choose Wikipedia
as the reference corpus when
computing the NPMI score, done by the Palmetto packages\footnote{http://palmetto.aksw.org} \citep{roder2015exploring}. 

\subsubsection{Backbones and Settings}
Our generalisation regularisation (\textbf{Greg}) framework can be easily applied to most of the NTMs. Here, we consider the following popular NTMs as our backbones: (i) Neural Variational Document Model (\textbf{NVDM}) \citep{miao2017discovering}, a pioneer NTM that applies Gaussian prior to $\bm{z}$. (ii) LDA with Products of Experts (\textbf{ProdLDA}) \citep{srivastava2017autoencoding}, an NTM that replaces the mixture of multinomials in LDA  with the product of experts. (iii) Neural Topic Model with
Covariates, Supervision, and Sparsity (\textbf{SCHOLAR}) \citep{card2017neural}, an NTM which applies logistic normal prior and incorporates metadata. (iv) Contrastive Neural Topic Model (\textbf{CLNTM}) \citep{nguyen2021contrastive}, a recent NTM that uses a contrastive learning framework to regularise the document representation. We follow the default settings for all these models, except the learning rates are fine-tuned to fit our own datasets. As for Greg, different DAs can be used; we use ``Highest to Similar'' in Table \ref{DA_Summary} throughout our experiments where the top 20 similar words are considered for replacement. The justification for this choice is described in Section \ref{sec:da_ablation}. As for hyperparameters of Greg, we set $\gamma$ as 300, $\beta$ as 0.5 for all experiments; As for the Sinkhorn algorithm \citep{cuturi2013sinkhorn}, we fix $\lambda$ as 100, the maximum number of iterations as 5,000 and the stop threshold as 0.005; As for the OT distance, we leverage the function with default settings in the POT package\footnote{https://pythonot.github.io/index.html} \citep{bonneel2011displacement} for the calculation.

\begin{table}[t]
  \caption{One Source (20News) to Different Targets K=50}
  \label{20News_Rest_k50}
  \centering
  \resizebox{0.5\textwidth}{!}{
  \begin{tabular}{ccc|cccc}
    \toprule
    \multirow{3}{*}{(\%,$\uparrow$)} & \multirow{3}{*}{Method}
    & \multicolumn{1}{c}{Source} & \multicolumn{4}{c}{Target}\\
    \cmidrule(l){3-3} \cmidrule(l){4-7} 
    & & 20News & Webs & TMN & DBpedia & R8 \\
    \midrule
    \multirow{9}{*}{CA} 
    & NVDM              & 40.0±0.5 & 39.0±0.8 & 38.8±0.5 & 31.2±0.6 & 73.2±0.6\\
    & + Greg            & \underline{42.0±0.6} & \underline{63.7±0.3} & \underline{60.2±0.4} & \underline{55.8±0.4} & \underline{80.6±0.5}\\
    \cmidrule(l){2-7}
    & PLDA               & 35.5±1.0  & 25.7±0.5   & 32.0±0.6 & 14.9±0.8 & 59.8±0.7\\
    & + Greg             & \underline{37.8±1.0} & \underline{31.6±0.4} & \underline{40.0±0.6} & \underline{19.0±0.4} & \underline{61.2±1.1}\\
    \cmidrule(l){2-7}
    & SCHOLAR          & \underline{53.5±1.0} & 56.0±0.7  & 51.8±0.3 & 50.9±1.2 & \underline{78.7±1.5}\\
    & + Greg           & 51.7±0.6 & \underline{59.6±1.9}   & \underline{57.8±1.8} & \underline{53.3±1.3} & 77.2±0.5\\
    \cmidrule(l){2-7}
    & CLNTM             & 48.5±0.9 & 46.5±1.7  & 43.8±1.1 & 42.0±1.4 & 74.9±1.9\\
    & + Greg            & \underline{48.7±0.7} & \underline{54.5±2.2} & \underline{54.0±1.6} & \underline{47.8±0.9} & \underline{76.7±1.3}\\
    \midrule
    \multirow{8}{*}{TP} 
    & NVDM              & 18.7±0.2 & 28.6±0.7 & 32.2±0.4 & 19.7±0.7 & 62.1±2.1\\
    & + Greg            & \underline{18.8±0.4} & \underline{35.8±1.5}   & \underline{40.1±1.5} & \underline{26.3±0.9} & \underline{63.7±1.8}\\
    \cmidrule(l){2-7}
    & PLDA              & 32.4±1.1  & 24.7±0.5   & 26.9±0.2 & 13.0±0.3 & 56.4±0.3\\
    & + Greg            & \underline{33.7±0.7} & \underline{25.7±0.8}   & \underline{29.5±0.8} & \underline{14.8±0.4} & \underline{57.7±0.8}\\
    \cmidrule(l){2-7}
    & SCHOLAR          & \underline{51.9±1.6} & 37.0±2.2  & 38.6±1.6 & 22.7±0.6 & \underline{61.2±1.1}\\
    & + Greg           & 48.3±0.9 & \underline{41.4±4.0}   & \underline{45.6±4.1} & \underline{23.1±2.6} & 61.0±1.2 \\
    \cmidrule(l){2-7}
    & CLNTM             & \underline{49.0±1.3} & 31.8±2.8  & 37.0±1.9 & 20.8±1.5 & 63.6±2.6\\
    & + Greg            & 36.8±2.0 & \underline{34.3±1.7}   & \underline{41.9±2.5} & \underline{21.1±1.9} & 63.6±4.7\\
    \midrule
    \multirow{8}{*}{TN} 
    & NVDM              & \underline{14.4±0.3} & 6.3±0.3 & 3.8±0.1 & 9.0±0.5 & 13.3±1.1 \\
    & + Greg            & 14.2±0.3  & \underline{12.7±0.8} & \underline{9.6±0.8} & \underline{15.7±0.5} & \underline{15.5±0.7}\\
    \cmidrule(l){2-7}
    & PLDA              & 21.3±0.7  & 3.6±0.2   & 1.6±0.1 & 4.0±0.2 & 7.7±0.3\\
    & + Greg            & \underline{23.8±0.6} & \underline{4.3±0.3}   & \underline{2.7±0.4} & \underline{5.1±0.3} & \underline{8.4±0.2}\\
    \cmidrule(l){2-7}
    & SCHOLAR          & \underline{45.3±1.1} & 16.2±0.9  & 11.3±1.1 & 15.1±0.8 & \underline{15.2±0.9}\\
    & + Greg           & 44.1±0.6 & \underline{19.5±2.7}   & \underline{19.3±3.2} & \underline{15.6±2.5} & 13.7±1.8\\
    \cmidrule(l){2-7}
    & CLNTM             & \underline{41.1±0.9} & 10.1±1.7  & 7.9±1.0 & 11.3±0.9 & 15.3±2.4\\
    & + Greg            & 36.3±2.0 & \underline{14.1±1.8}   & \underline{14.6±2.6} & \underline{12.4±1.5} & \underline{15.4±4.5}\\
  \bottomrule
\end{tabular}
}
\end{table}

\subsection{Results and Analysis}
\subsubsection{One Source to Different Targets} \label{sec:one_targets}
Here, we set 20News as the source corpus and the other datasets as the target corpora for our experiments. The quality of the topical representation is measured by CA, TP and TN. The results for $K=50$ are illustrated in Table \ref{20News_Rest_k50}, where the larger value between backbone and backbone with Greg under each setting is underlined. We have the following remarks from the results: (i) When applying Greg to different models, the CA, TP and TN on different target corpora are significantly improved in most cases. For example, by integrating NVDM with Greg, we improve CA from 39\% to 63.7\%, 38.8\% to 60.2\%, 31.2\% to 55.8\% and 73.2\% to 80.6\% on average for targets Webs, TMN, DBpedia and R8, respectively. Similarly, a large improvement can also be obtained for TP and TN after applying Greg. (ii) The performance on the source corpus is also improved when applied to different backbones under most settings. For example, when applying Greg to PLDA, the performance is increased from 35.5\% to 37.8\%, 32.4\% to 33.7\% and 21.3\% to 23.8\% on average for the source 20News in terms of CA, TP and TN, respectively. Similar observations can be obtained from the results in Table S1-S2, which illustrates the results for different settings of $K$. Overall, the results show that our approach effectively generalises the neural topical representation across corpora in our experiments.

\begin{table}[!t]
  \caption{Different Sources to One Target (TMN) K = 50}
  \label{target_TMN_k50}
  \centering
  \resizebox{0.5\textwidth}{!}{
  \begin{tabular}{cccccc|c}
    \toprule
    \multirow{3}{*}{(\%,$\uparrow$)} & \multirow{3}{*}{Method}
    & \multicolumn{4}{c}{$\text{Target}$ (TMN)} & $\text{Original}$ \\
     \cmidrule(l){3-6} \cmidrule(l){7-7} 
    & & 20News\_T & Webs\_T & DBpedia\_T & R8\_T & TMN \\
    \midrule
    \multirow{9}{*}{CA} 
    & NVDM              & 38.4±0.3 & 44.4±0.3 & 39.5±1.8 & 33.5±0.6 & 61.9±0.8\\
    & + Greg            & \underline{59.6±0.4} & \underline{63.4±0.6} & \underline{59.4±0.6} & \underline{43.3±0.7} & \underline{67.1±1.1}\\
    \cmidrule(l){2-7}
    & PLDA                & 31.7±0.1 & 43.1±0.6 & 37.4±0.6 & 26.8±0.2 & 62.7±0.8\\
    & + Greg              & \underline{40.7±0.5} & \underline{48.8±0.8} & \underline{42.0±0.3} & \underline{27.3±0.1} & \underline{64.5±1.1}\\
    \cmidrule(l){2-7}
    & SCHOLAR          & 51.3±0.6 & 49.3±0.8 & 56.8±0.7 & 43.8±0.7 & 73.0±0.5\\
    & + Greg           & \underline{58.5±1.8} & \underline{63.6±1.0} & \underline{60.9±1.4} & \underline{45.9±0.7} & \underline{82.6±0.5}\\
    \cmidrule(l){2-7}
    & CLNTM            & 44.8±1.3 & 45.4±1.3 & 53.8±0.8 & 42.2±0.8 & 72.6±0.7\\
    & + Greg           &  \underline{54.1±2.2} & \underline{64.1±1.3} & \underline{60.6±1.5} & \underline{45.1±0.8} & \underline{82.0±0.4}\\
    \midrule
    \multirow{9}{*}{TP} 
    & NVDM              & 32.0±0.3 & 31.5±0.2 & 30.8±0.7 & 29.4±0.7 & 36.0±0.6\\
    & + Greg            & \underline{40.0±0.7} & \underline{38.9±0.6} & \underline{38.0±1.4} & \underline{33.7±1.1} & \underline{38.2±1.1}\\
    \cmidrule(l){2-7}
    & PLDA                & 27.0±0.2 & 37.4±0.6 & 30.6±0.4 & 26.3±0.0 & 62.5±0.9\\ 
    & + Greg              & \underline{30.5±0.7} & \underline{41.7±0.6} & \underline{32.3±0.6} & \underline{26.5±0.1} & \underline{63.8±1.1}\\
    \cmidrule(l){2-7}
    & SCHOLAR          & 37.8±5.3 & 35.5±3.4 & 49.2±2.3 & 29.3±1.2 & 58.5±1.6\\
    & + Greg           & \underline{48.7±1.9} & \underline{55.8±2.4} & \underline{53.4±1.3} & \underline{33.8±1.2} & \underline{81.0±0.8} \\
    \cmidrule(l){2-7}
    & CLNTM            & 38.2±3.3 & 33.1±1.1 & 45.9±2.0 & 34.1±1.5 & 59.7±2.5\\
    & + Greg           & \underline{42.2±3.9} & \underline{58.0±1.3} & \underline{53.9±2.0} & 34.1±2.5 & \underline{79.7±1.1}\\
    \midrule
    \multirow{9}{*}{TN} 
    & NVDM             & 3.5±0.1 & 3.6±0.1 & 3.0±0.2 & 2.3±0.2 & 6.5±0.3\\
    & + Greg           & \underline{9.7±0.4} & \underline{8.9±0.2} & \underline{8.1±1.0} & \underline{4.7±0.3} & \underline{8.2±0.4} \\
    \cmidrule(l){2-7}
    & PLDA                & 1.5±0.1 & 6.8±0.4 & 3.1±0.2 & 1.1±0.0 & 23.6±0.3\\
    & + Greg              & \underline{3.3±0.2} & \underline{9.6±0.6} & \underline{4.5±0.3} & 1.1±0.0 & \underline{25.1±0.6}\\
    \cmidrule(l){2-7}
    & SCHOLAR          & 12.8±1.7 & 12.0±2.1 & 18.2±1.0 & 3.4±0.8 & 37.8±1.2\\
    & + Greg           & \underline{20.5±1.4} & \underline{23.1±2.4} & \underline{24.3±1.2} & \underline{7.2±0.8} & \underline{44.5±0.4}\\
    \cmidrule(l){2-7}
    & CLNTM            & 8.5±2.1 & 6.6±1.9 & 17.0±0.9 & 6.7±0.7 & 39.1±1.8\\
    & + Greg           & \underline{14.4±3.5} & \underline{24.2±0.5} & \underline{23.8±1.5} & \underline{8.1±1.5} & \underline{43.4±1.1}\\
  \bottomrule
\end{tabular}
}
\end{table}

\begin{table}[!t]
  \caption{Different Sources to One Target (R8) K = 50}
  \label{target_R8_k50}
  \centering
  \resizebox{0.5\textwidth}{!}{
  \begin{tabular}{cccccc|c}
    \toprule
    \multirow{3}{*}{(\%,$\uparrow$)} & \multirow{3}{*}{Method}
    & \multicolumn{4}{c}{$\text{Target}$ (R8)} & $\text{Original}$ \\
    \cmidrule(l){3-6} \cmidrule(l){7-7} 
    & & 20News\_R & Webs\_R & TMN\_R & DBpedia\_R & R8 \\
    \midrule
    \multirow{9}{*}{CA} 
    & NVDM              & 72.6±0.3 & 73.8±0.9 & 78.6±0.6 & 72.2±0.6 & 87.6±0.6\\
    & + Greg            & \underline{80.8±1.4} & \underline{80.1±1.0} & \underline{83.7±0.6} & \underline{73.8±0.9} & \underline{88.7±0.4}\\
    \cmidrule(l){2-7}
    & PLDA                & 59.7±0.3 & 63.5±0.9 & 74.9±1.2 & 63.6±0.6 & 79.7±0.5\\
    & + Greg              & \underline{63.4±0.7} & \underline{64.4±0.8} & \underline{75.9±0.7} &\underline{63.9±0.6} & \underline{81.5±0.6}\\
    \cmidrule(l){2-7}
    & SCHOLAR          & \underline{77.8±1.6} & 76.8±1.1 & 76.5±1.5 & \underline{76.0±0.2} & 91.4±0.4\\
    & + Greg           & 75.2±1.5 & \underline{77.6±0.7} & \underline{84.2±0.9} & 75.1±0.8 & \underline{93.3±0.4}\\
    \cmidrule(l){2-7}
    & CLNTM            & 74.0±1.6 & 75.9±1.3 & 76.6±1.0 & 73.5±0.6 & 92.1±0.4\\
    & + Greg           & \underline{75.0±1.5} & \underline{77.5±1.1} & \underline{81.9±2.3} & \underline{75.5±0.6} & \underline{92.9±0.3}\\
    \midrule
    \multirow{9}{*}{TP} 
    & NVDM              & 61.2±0.5 & 62.2±1.1 & 61.0±1.5 & \underline{61.9±2.0} & 70.2±1.6\\
    & + Greg            & \underline{65.1±2.9} & \underline{64.8±2.6} & \underline{64.7±1.5} & 61.8±2.2 & \underline{70.3±1.7}\\
    \cmidrule(l){2-7}
    & PLDA                & 56.7±0.4 & \underline{58.2±0.7} & 64.5±2.6 & 61.1±0.7 & 74.7±0.2\\ 
    & + Greg              & \underline{57.8±1.1} & 56.7±0.7 & \underline{68.5±2.3} & \underline{62.1±0.9} & \underline{77.2±0.9}\\
    \cmidrule(l){2-7}
    & SCHOLAR          & 61.6±1.5 & 64.3±0.6 & 58.8±3.4 & \underline{64.1±0.6} & \underline{91.1±0.6}\\
    & + Greg           & \underline{63.0±2.0} & \underline{64.4±1.4} & \underline{70.6±5.3} & 63.9±2.1 & 88.9±0.9\\
    \cmidrule(l){2-7}
    & CLNTM            & \underline{63.0±2.2} & \underline{65.0±2.1} & 58.8±0.8 & \underline{64.8±1.4} & \underline{91.1±1.7}\\
    & + Greg           & 62.8±2.1 & 64.3±1.4 & \underline{68.9±5.4} & 62.5±2.3 & 90.0±1.5\\
    \midrule
    \multirow{9}{*}{TN} 
    & NVDM             & 12.9±0.5 & 13.4±0.8 & 13.2±0.6 & 12.6±0.9 & \underline{22.0±0.7}\\
    & + Greg           & \underline{17.5±1.5} & \underline{17.5±2.3} & \underline{16.9±1.4} & \underline{13.1±1.1} & 21.7±0.7\\
    \cmidrule(l){2-7}
    & PLDA                & 7.8±0.3 & \underline{10.6±0.9} & 20.0±1.7 & 11.8±0.2 & 24.0±0.4\\
    & + Greg              & \underline{8.7±0.4} & 10.5±0.7 & \underline{23.2±2.5} & \underline{12.4±0.7} & \underline{26.4±0.7}\\
    \cmidrule(l){2-7}
    & SCHOLAR          & 14.7±1.2 & 20.6±1.9 & 12.4±2.4 & \underline{20.2±0.7} & 43.2±0.6\\
    & + Greg           & \underline{16.2±2.2} & \underline{20.8±1.2} & \underline{35.4±5.0} & 18.9±1.2 & \underline{44.8±1.2}\\
    \cmidrule(l){2-7}
    & CLNTM            & 15.0±1.2 & 18.8±1.9 & 13.4±2.0 & \underline{20.3±1.2} & 43.7±1.4\\
    & + Greg           & 15.0±1.6 & \underline{19.7±1.7} & \underline{34.2±4.8} & 18.0±0.8 & \underline{45.3±1.8}\\
  \bottomrule
\end{tabular}
}
\end{table}

\begin{table}[!ht]
  \caption{Target as Noisy Corpus K = 50}
  \label{noisy_corpus_k50}
  \centering
  \resizebox{0.5\textwidth}{!}{
  \begin{tabular}{cc|ccccc}
    \toprule
    (\%,$\uparrow$) & Method & 20News\_N & Webs\_N & TMN\_N & DBpedia\_N & R8\_N\\
    \midrule
    \multirow{8}{*}{CA} 
    & NVDM              & 25.8±0.2 & 49.7±0.9 & 51.3±0.5 & 58.5±1.0 & 80.3±0.4\\
    & + Greg            & \underline{28.0±0.4} & \underline{60.1±0.8} & \underline{57.5±0.9} & \underline{66.1±0.8} & \underline{83.5±1.0}\\
    \cmidrule(l){2-7}
    & PLDA                & 23.1±0.6 & 52.2±0.9 & 52.1±0.6 & 55.3±2.1 & 70.6±0.4\\
    & + Greg              & \underline{25.3±0.6} & \underline{55.4±0.4} & \underline{54.1±0.7} & \underline{58.2±1.1} & \underline{73.0±0.9}\\
    \cmidrule(l){2-7}
    & SCHOLAR             & \underline{43.2±0.7} & 73.6±2.9 & 66.1±0.9 & 82.7±1.4 & 87.3±0.9\\
    & + Greg              & 42.3±2.1 & \underline{86.0±0.7} & \underline{77.5±0.3} & \underline{84.1±1.4} & \underline{88.4±0.5}\\
    \cmidrule(l){2-7}
    & CLNTM               & \underline{39.8±1.3} & 70.7±1.3 & 66.1±0.7 & 70.0±1.6 & 87.3±0.6\\
    & + Greg              & 39.5±1.8 & \underline{87.1±0.6} & \underline{77.2±0.7} & \underline{82.2±1.0} & \underline{87.7±0.8}\\
    \midrule
    \multirow{8}{*}{TP} 
    & NVDM              & 14.4±0.4 & 30.4±0.5 & 33.1±0.8 & 24.7±1.2 & 66.3±1.1\\
    & + Greg            & \underline{14.7±0.5} & \underline{33.0±0.8} & \underline{35.7±1.0} & \underline{26.6±1.3} & \underline{66.8±1.3}\\
    \cmidrule(l){2-7}
    & PLDA                & 21.6±0.5 & 50.3±0.6 & 50.8±0.8 & 58.2±0.6 & 64.7±0.6\\
    & + Greg              & \underline{23.3±0.4} & \underline{53.7±0.4} & \underline{52.6±0.6} & \underline{62.0±1.1} & \underline{66.8±1.0}\\
    \cmidrule(l){2-7}
    & SCHOLAR          & \underline{39.5±1.0} & 46.6±2.7 & 54.0±1.5 & 74.2±1.9 & \underline{83.7±1.2}\\
    & + Greg           & 37.7±1.3 & \underline{83.1±1.7} & \underline{76.7±0.9} & \underline{86.4±0.8} & 83.0±0.6\\
    \cmidrule(l){2-7}
    & CLNTM            & \underline{37.4±1.1} & 46.7±2.4 & 54.6±2.9 & 59.0±4.0 & \underline{80.9±1.9}\\
    & + Greg           & 30.2±3.8 & \underline{84.6±1.2} & \underline{74.9±1.1} & \underline{81.7±1.3} & 80.4±2.1\\
    \midrule
    \multirow{8}{*}{TN} 
    & NVDM              & 8.8±0.2 & 7.3±0.1 & 4.5±0.2 & 14.1±0.7 & 16.5±0.6\\
    & + Greg            & \underline{9.0±0.4} & \underline{9.1±0.2} & \underline{6.3±0.4} & \underline{17.3±0.8} & \underline{17.3±1.0}\\
    \cmidrule(l){2-7}
    & PLDA                & 10.7±0.4 & 18.3±0.4 & 14.0±0.1 & 35.8±0.5 & 15.1±0.4\\
    & + Greg              & \underline{12.7±0.4} & \underline{21.1±0.5} & \underline{15.4±0.4} & \underline{40.0±0.7} & \underline{16.6±0.3}\\
    \cmidrule(l){2-7}
    & SCHOLAR          & 31.6±0.9 & 29.6±3.5 & 31.7±1.8 & \underline{69.1±1.4} & 36.4±1.3\\
    & + Greg           & \underline{32.0±1.1} & \underline{53.2±1.4} & \underline{39.1±0.4} & 68.5±1.1 & \underline{37.3±0.8}\\
    \cmidrule(l){2-7}
    & CLNTM            & \underline{30.4±1.4} & 28.8±2.0 & 33.3±1.9 & 54.5±3.2 & 34.3±1.3\\
    & + Greg           & 26.1±2.3 & \underline{55.1±1.4} & \underline{38.4±1.2} & \underline{64.9±0.9} & \underline{36.2±1.8}\\
  \bottomrule
\end{tabular}
}
\end{table}

\begin{table}[!ht]
  \caption{Source Corpus Performance K = 50}
  \label{pure_source_k50}
  \centering
  \resizebox{0.5\textwidth}{!}{
  \begin{tabular}{cc|ccccc}
    \toprule
    (\%,$\uparrow$) & Method & 20News & Webs & TMN & DBpedia & R8 \\
    \midrule
    \multirow{8}{*}{CA} 
    & NVDM              & 40.1±0.2 & 60.1±0.8 & 61.9±0.8 & 74.5±0.6 & 87.6±0.6 \\
    & + Greg            & \underline{41.9±0.4} & \underline{69.5±0.4} & \underline{67.1±1.1} & \underline{79.3±0.5} & \underline{88.7±0.4}\\
    \cmidrule(l){2-7}
    & PLDA                & 35.2±0.8 & 62.0±1.3 & 62.7±0.8 & 69.5±1.1 & 79.7±0.5\\
    & + Greg              & \underline{37.7±0.5} & \underline{65.3±1.1} & \underline{64.5±1.1} & \underline{72.1±1.9} & \underline{81.5±0.6}\\
    \cmidrule(l){2-7}
    & SCHOLAR               & \underline{56.3±0.8} & 81.2±2.2 & 73.0±0.5 & 89.1±1.2 & 91.4±0.4 \\
    & + Greg           & 54.1±2.2 & \underline{88.1±0.9} & \underline{82.6±0.5} & \underline{91.0±1.1} & \underline{93.3±0.4}\\
    \cmidrule(l){2-7}
    & CLNTM               & \underline{55.6±1.0} & 77.9±2.0 & 72.6±0.7 & 76.2±1.3 & 92.1±0.4 \\
    & + Greg           & 53.0±1.3 & \underline{91.1±0.7} & \underline{82.0±0.4} & \underline{88.8±0.7} & \underline{92.9±0.3}\\
    \midrule
    \multirow{8}{*}{TP} 
    & NVDM              & \underline{19.7±0.4} & 32.2±0.7 & 36.0±0.6 & 30.0±1.2 & 70.2±1.6\\
    & + Greg            & 19.1±0.5 & \underline{34.4±0.7} & \underline{38.2±1.1} & \underline{31.5±1.3} & \underline{70.3±1.7}\\
    \cmidrule(l){2-7}
    & PLDA                & 33.6±0.7 & 60.5±1.3 & 62.5±0.9 & 74.5±1.1 & 74.7±0.2 \\
    & + Greg              & \underline{35.8±0.5} & \underline{63.9±0.9} & \underline{63.8±1.1} & \underline{77.0±0.7} & \underline{77.2±0.9}\\
    \cmidrule(l){2-7}
    & SCHOLAR               & \underline{54.6±0.8} & 51.6±3.1 & 58.5±1.6 & 80.1±3.0 & \underline{91.1±0.6}\\
    & + Greg           & 51.0±1.8 & \underline{85.4±1.7} & \underline{81.0±0.8} & \underline{91.5±1.1} & 88.9±0.9\\
    \cmidrule(l){2-7}
    & CLNTM               & \underline{57.5±1.0} & 51.2±3.3 & 59.7±2.5 & 63.5±4.2 & \underline{91.1±1.7} \\
    & + Greg           &  44.5±5.0 & \underline{88.8±0.9} & \underline{79.7±1.1} & \underline{88.2±1.4} & 90.0±1.5 \\
    \midrule
    \multirow{8}{*}{TN} 
    & NVDM              & \underline{14.2±0.4} & 8.3±0.4 & 6.5±0.3 & 20.3±1.1 & \underline{22.0±0.7} \\
    & + Greg            &  14.0±0.5 & \underline{10.2±0.3} & \underline{8.2±0.4} & \underline{23.3±1.1} & 21.7±0.7\\
    \cmidrule(l){2-7}
    & PLDA                & 22.3±0.3 & 27.4±1.0 & 23.6±0.3 & 54.2±0.4 & 24.0±0.4 \\
    & + Greg              & \underline{25.4±0.4} & \underline{30.7±0.9} & \underline{25.1±0.6} & \underline{57.2±0.4} & \underline{26.4±0.7}\\
    \cmidrule(l){2-7}
    & SCHOLAR               & \underline{48.1±0.7} & 34.2±3.6 & 37.8±1.2 & \underline{78.4±1.6} & 43.2±0.6 \\
    & + Greg           & 46.7±1.7 & \underline{55.8±1.5} & \underline{44.5±0.4} & 75.3±1.3 & \underline{44.8±1.2}\\
    \cmidrule(l){2-7}
    & CLNTM               & \underline{49.1±0.3} & 34.7±2.5 & 39.1±1.8 & 63.1±3.4 & 43.7±1.4\\
    & + Greg           & 42.5±3.4 & \underline{60.7±1.2} & \underline{43.4±1.1} & \underline{72.6±0.7} & \underline{45.3±1.8}\\
  \bottomrule
\end{tabular}
}
\end{table}

\subsubsection{Different Sources to One Target}\label{sec:diff_sources}
Here, we fix the target and use different source datasets to further investigate NTM's generalisation ability. For the target, we use one short document corpus TMN and one long document corpus R8, respectively. Then the rest datasets are set as the sources, respectively. The results for the target as TMN and R8 when $K$ equals 50 are illustrated in Table \ref{target_TMN_k50} and Table \ref{target_R8_k50}, respectively. Notably, ``\textbf{20News\_T}'' indicates the evaluation is conducted on target corpus TMN where the model is trained on source corpus 20News; ``\textbf{20News\_R}'' indicates the evaluation is conducted on target corpus R8 where the model is trained on source corpus 20News; ``\textbf{Original}'' indicates the model is both trained and evaluated on the same corpus. Based on these results, we summarise the following observations: (i) We significantly improve documents' topical representation in the target corpus when different source datasets are used. For example, from Table \ref{target_TMN_k50}, in terms of CA, by integrating Greg with NVDM, the performance in the target TMN is increased from 38.4\% to 59.6\%, 44.4\% to 63.4\%, 39.5\% to 59.4\% and 33.5\% to 43.3\% when 20News, Webs, DBpedia and R8 are used as the source, respectively. (ii) For some cases, Greg even generalises the topical representation to the target from the source with better performance than the model trained on the target corpus directly (i.e. the value in the ``Original'' columns). For example, from Table \ref{target_TMN_k50}, training NVDM on TMN can achieve a CA of 61.9\%. While the performance can be even 63.4\% when using NVDM with Greg training on Webs without touching the data in TMN, demonstrating our framework's generalisation power. (iii) Similar observations can be obtained for both TMN and R8 as the targets, as well as in different settings of $K$ (Table S3-S6).

\subsubsection{Target as Noisy Corpus}\label{sec:noisy_targets}
Here, we challenge the models with noisy datasets as the target, where the model is trained on original source
datasets but evaluated on their noisy versions (e.g. \textbf{Dataset\_N}). The experimental results on the noisy targets at $K=50$ are illustrated in Table \ref{noisy_corpus_k50}. As for other settings, including $K=20$ and $K=100$, are illustrated in Table S7-S8. From this set of experiments, we can observe that (i) Greg continues showing its benefits in generalising neural topical representation when the target domain contains noise in most settings, demonstrating improved robustness at the same time. (ii) It can be noticed that Greg causes the performance drop when applying to CLNTM on 20News, the potential reason is that CLNTM uses a contrastive learning approach to regularise the topical representation in addition to our Greg, thus the balance between 2 regulariser is harder to be achieved during the training.

\subsubsection{Source Corpus Performance}
We illustrate the performance on the original corpus when applying Greg to different NTMs in Table \ref{pure_source_k50}. The experimental results for other settings of $K$ are illustrated in Table S9-S10. Based on these results, it can be observed that (i) Greg can improve the topical representation quality of the source documents at the same time for short documents (i.e., Webs, TMN and DBpedia) under most settings. (ii) For long document corpora such as 20News and R8, there are a few cases where Greg leads to a performance drop on the source corpus, while we believe that the benefits of Greg are significant and the performance can be improved further for particular datasets with hyperparameter tunning.

\begin{table}[!t]
  \caption{P-Value of Paired T-Test on Target Corpus ($\alpha = 0.05$)}
  \vspace{1mm}
  \label{significant_target}
  \centering
  \begin{tabular}{ccccc}
    \toprule
    \multirow{3}{*}{Metric} & \multicolumn{4}{c}{Target}\\
    \cmidrule(l){2-5} 
    & Webs & TMN & DBpedia & R8 \\
    \midrule
    CA & \underline{3.66e-12} & \underline{6.49e-70} & \underline{4.70e-09} & \underline{1.25e-25}\\
    TP & \underline{4.88e-07} & \underline{1.68e-43} & \underline{7.35e-05} & \underline{9.71e-11}\\
    TN & \underline{1.31e-09} & \underline{5.95e-55} & \underline{1.87e-05} & \underline{1.23e-12}\\
  \bottomrule
\end{tabular}
\end{table}
\begin{table}[!t]
  \caption{P-Value of Paired T-Test on Noisy Corpus ($\alpha = 0.05$)}
  \vspace{1mm}
  \label{significant_noisy}
  \centering
  \begin{tabular}{cccccc}
    \toprule
    \multirow{3}{*}{Metric} & \multicolumn{5}{c}{Noisy Target}\\
    \cmidrule(l){2-6} 
    & Webs & TMN & DBpedia & 20News & R8 \\
    \midrule
    CA & \underline{1.42e-12} & \underline{3.79e-14} & \underline{1.21e-10} & 7.01e-02 & \underline{2.67e-09}\\
    TP & \underline{2.98e-07} & \underline{2.25e-10} & \underline{9.90e-07} & 7.45e-01 & 4.85e-01\\
    TN & \underline{1.11e-07} & \underline{8.03e-10} & \underline{4.44e-07} & 6.37e-01 & \underline{1.30e-04}\\
  \bottomrule
\end{tabular}
\end{table}
\begin{table}[!t]
  \caption{P-Value of Paired T-Test on Source Corpus ($\alpha = 0.05$)}
  \vspace{1mm}
  \label{significant_source}
  \centering
  \begin{tabular}{cccccc}
    \toprule
    \multirow{3}{*}{Metric} & \multicolumn{5}{c}{Source}\\
    \cmidrule(l){2-6} 
    & Webs & TMN & DBpedia & 20News & R8 \\
    \midrule
    CA & \underline{2.24e-12} & \underline{5.12e-14} & \underline{5.53e-11} & 3.42e-01 & \underline{2.96e-16}\\
    TP & \underline{1.67e-07} & \underline{1.43e-09} & \underline{1.23e-06} & \underline{3.10e-05} & 5.61e-01\\
    TN & \underline{8.88e-08} & \underline{3.54e-08} & \underline{3.62e-06} & \underline{2.23e-02} & \underline{1.20e-03}\\
  \bottomrule
\end{tabular}
\end{table}

\subsubsection{Significance Test}
We conduct paired t-tests across our previous experimental results to demonstrate Greg's performance across different scenarios from a general view. We collect paired differences in performance metrics between the original model and the model incorporating Greg across various datasets under the following settings: (i) different target corpora (Table \ref{significant_target}), (ii) noisy versions of the source corpora (Table \ref{significant_noisy}), and (iii) the original source corpora (Table \ref{significant_source}). We set the significance level (i.e., $\alpha$) as 0.05 for all paired t-tests. The P-value lower than $\alpha$ is underlined in tables, indicating a significant difference between the model with and without Greg. We have the following observations based on the results: (i) From Table \ref{significant_target}, the benefit of Greg that improves the performance of different target corpora is significant for all targets. (ii) From Table \ref{significant_noisy}, when the target is a noisy corpus, the improvement by Greg is significant for all short corpora (e.g., Webs, TMN and DBpedia). For the long-document noisy target such as 20News, Greg shows comparable performance with the original model. (iii) From Table \ref{significant_source}, the improvement to source corpus performance by Greg is significant for most settings.

\begin{table*}[ht]
  \caption{Effect of Different DA on Topical Representation}
  \label{DA_effect}
  \centering
  \resizebox{\textwidth}{!}{
  \begin{tabular}{ccccc|ccc|ccc}
    \toprule
    \multirow{3}{*}{Dataset} & \multirow{3}{*}{Method}
    & \multicolumn{3}{c}{CD (\%, $\downarrow$)} & \multicolumn{3}{c}{HD (\%, $\downarrow$)} & \multicolumn{3}{c}{TopicalOT (\%, $\downarrow$)}\\
    \cmidrule(l){3-5} \cmidrule(l){6-8} \cmidrule(l){9-11}
    & & LDA & NVDM & \normalfont{NVDM + Greg} & LDA & NVDM & \normalfont{NVDM + Greg} & LDA & NVDM & \normalfont{NVDM + Greg}\\
    \midrule
    \multirow{8}{*}{20News} 
    & Random Drop             & 13.3±0.4 & 9.8±0.1  & \underline{7.3±0.1}  & 33.3±0.4 & 15.8±0.1 & \underline{13.6±0.1} & 10.8±0.2 & 8.9±0.1  & \underline{3.8±0.1}\\
    & Random Insertion        & 14.2±0.4 & 9.7±0.1  & \underline{7.0±0.1}  & 38.5±0.4 & 15.7±0.1 & \underline{13.2±0.1} & 13.2±0.2 & 8.8±0.0  & \underline{3.6±0.1}\\
    \cmidrule(l){2-11}
    & Random to Similar       & 15.8±0.3 & 11.1±0.1 & \underline{7.2±0.1}  & 37.6±0.3 & 16.9±0.1 & \underline{13.6±0.1} & 12.5±0.2 & 9.5±0.0  & \underline{3.7±0.1}\\
    & Highest to Similar      & 17.6±0.5 & 11.3±0.1 & \underline{7.1±0.1}  & 38.8±0.4 & 17.0±0.1 & \underline{13.4±0.1} & 12.7±0.3 & 9.5±0.0  & \underline{3.7±0.1}\\
    & Lowest to Similar       & 17.6±0.1 & 10.2±0.1 & \underline{7.4±0.1}  & 39.6±0.3 & 16.1±0.1 & \underline{13.6±0.1} & 14.2±0.2 & 9.0±0.0  & \underline{3.8±0.1}\\
    \cmidrule(l){2-11}
    & Random to Dissimilar    & 26.3±1.6 & 20.7±1.1 & \underline{15.0±0.5} & 47.7±1.0 & 23.7±0.7 & \underline{19.9±0.4} & 18.4±0.4 & 13.2±0.4 & \underline{5.5±0.1}\\
    & Highest to Dissimilar   & 32.4±2.0 & 21.2±1.0 & \underline{15.0±0.4} & 51.5±1.1 & 23.9±0.6 & \underline{19.9±0.3} & 20.8±0.5 & 13.4±0.4 & \underline{5.5±0.1}\\
    & Lowest to Dissimilar    & 20.4±1.1 & 19.5±1.2 & \underline{14.7±0.4} & 44.0±0.9 & 22.8±0.7 & \underline{19.6±0.3} & 16.0±0.4 & 12.8±0.4 & \underline{5.4±0.1}\\
   \midrule
    \multirow{8}{*}{Webs} 
    & Random Drop             & 5.9±0.2 & 10.0±0.1 & \underline{5.6±0.1} & 20.2±0.3 & 16.5±0.0 & \underline{12.3±0.1} & 8.7±0.1  & 8.8±0.2  & \underline{3.1±0.1}\\
    & Random Insertion        & 4.8±0.2 & 9.4±0.1  & \underline{4.7±0.1} & 23.6±0.3 & 15.8±0.1 & \underline{11.0±0.1} & 9.3±0.1  & 8.4±0.1  & \underline{2.8±0.1}\\
    \cmidrule(l){2-11}
    & Random to Similar       & 5.6±0.4 & 10.8±0.1 & \underline{5.0±0.1} & 21.4±0.5 & 17.1±0.1 & \underline{11.4±0.1} & 8.4±0.2  & 9.0±0.2  & \underline{2.9±0.1}\\
    & Highest to Similar      & 5.8±0.4 & 10.3±0.2 & \underline{4.7±0.1} & 21.1±0.4 & 16.7±0.1 & \underline{11.2±0.1} & 8.3±0.2  & 8.9±0.2  & \underline{2.9±0.1}\\
    & Lowest to Similar       & 4.8±0.3 & 11.0±0.1 & \underline{5.3±0.1} & 21.2±0.4 & 17.2±0.1 & \underline{11.7±0.1} & 8.2±0.1  & 9.1±0.2  & \underline{3.0±0.1}\\
    \cmidrule(l){2-11}
    & Random to Dissimilar    & 8.8±0.3 & 11.9±0.2 & \underline{6.4±0.1} & 29.6±0.4 & 18.1±0.2 & \underline{13.0±0.1} & 12.6±0.2 & 9.6±0.2  & \underline{3.3±0.1}\\
    & Highest to Dissimilar   & 8.4±0.3 & 11.3±0.1 & \underline{6.1±0.1} & 28.9±0.4 & 17.7±0.1 & \underline{12.8±0.1} & 12.5±0.2 & 9.4±0.1  & \underline{3.2±0.1}\\
    & Lowest to Dissimilar    & 7.7±0.4 & 11.9±0.3 & \underline{6.6±0.2} & 27.7±0.7 & 18.0±0.2 & \underline{13.2±0.2} & 11.5±0.3 & 9.6±0.1  & \underline{3.4±0.1}\\
  \bottomrule
\end{tabular}
}
\end{table*}

\subsection{Ablation Study and Hyperparameter Sensitivity}
\subsubsection{Effect of DA on Topical Representation}\label{DA_study}
In this section, we study the effect of different DAs on topical representations. 
As most topic models work with BOWs, we focus on the word-level DAs described in Table \ref{DA_Summary}. The effect of DAs on topical representation is measured by the distance between the topical representations of original documents and augmentations. Specifically, we use the trained model to infer the topical representation of the test documents and their augmentations; Then different distance metrics are applied to calculate the distance between the topical representations of a document and its augmentations. The choices of distance metrics include Cosine distance (CD), Hellinger distance (HD) and TopicalOT. The choice of models here includes LDA, NVDM and NVDM with Greg; Moreover, we train these models with $K=50$ on one long document corpus 20News and one short document corpus Webs; For the settings of DA, the augmentation strength is set as 0.5. For approaches based on word similarities, the number of top similar/dissimilar words considered for replacement is 20, where the Cosine similarity between GloVe word embeddings is applied to provide word similarity.

The results are illustrated in Table \ref{DA_effect}. We have the following observations from this study: (i) When more words are perturbed (e.g. in long documents of 20News), NTMs such as NVDM are more stable (i.e. lower distances obtained) to DAs than probabilistic topic models such as LDA. While in the case that fewer words are perturbed, such as in Webs, LDA is more stable than NVDM.  (ii) DAs that replace with similar words bring less effect than those that replace with dissimilar words. (iii) Interestingly, adding noise by random drop or insertion has a similar effect to replacing with similar words in our settings. These observations are cues of the intrinsic generalisation ability of NTMs, which is further enhanced by Greg in this work.

\begin{table}[!t]
  \caption{Effect of DA on Greg}
  \label{Ablation_da}
  \centering
  \resizebox{0.5\textwidth}{!}{
  \begin{tabular}{cccccc}
    \toprule
    \multirow{3}{*}{(\%,$\uparrow$)} & \multirow{3}{*}{Method}  
    & \multicolumn{4}{c}{Targets} \\
     \cmidrule(l){3-6} 
    & & Webs & TMN & DBpedia & R8 \\
    \midrule
    \multirow{4}{*}{CA} 
    & Random to Similar          & 62.7±0.4 & 58.3±0.2 & \underline{56.0±0.7} & 80.5±0.5 \\
    & Highest to Similar         & \underline{63.7±0.3} & \underline{60.2±0.4} & 55.8±0.4 & \underline{80.6±0.5}\\
    & Lowest to Similar          & 55.5±0.7 & 52.8±0.8 & 49.6±0.8 & 78.9±0.8 \\
    & Both to Similar            & 60.7±1.2 & 57.3±0.1 & 52.4±1.5 & 79.3±0.9 \\
    \midrule
    \multirow{4}{*}{TP} 
    & Random to Similar          & 34.7±1.0 & 39.6±1.6 & 25.5±0.9 & 63.5±1.8 \\
    & Highest to Similar         & \underline{35.8±1.5} & \underline{40.1±1.5} & \underline{26.3±0.9} & \underline{63.7±1.8}\\
    & Lowest to Similar           & 33.6±0.7 & 37.4±1.2 & 23.9±1.1 & 62.6±1.7 \\
    & Both to Similar             & 34.8±1.2 & 38.6±1.7 & 25.5±1.2 & 62.4±2.1 \\
    \midrule
    \multirow{4}{*}{TN} 
    & Random to Similar           & 12.5±1.0 & 9.1±0.7 & 15.6±0.6 & 15.3±1.0  \\
    & Highest to Similar          & \underline{12.7±0.8} & \underline{9.6±0.8} & \underline{15.7±0.5} & \underline{15.5±0.7}\\
    & Lowest to Similar           & 10.8±0.3 & 7.4±0.5 & 13.8±0.5 & 14.5±1.3 \\
    & Both to Similar             & 11.9±0.9 & 8.4±0.9 & 15.0±0.9 & 14.9±0.9 \\
  \bottomrule
\end{tabular}
}
\end{table}

\subsubsection{Effect of Positive DA on Greg}\label{sec:da_ablation}
Different DAs can be applied for Greg as long as it creates positive (e.g. similar) documents, which shows the flexibility of our framework. Here, we demonstrate our choice of DA by applying different positive DAs to Greg. Their resulting topical representation quality of target corpora is illustrated in Table \ref{Ablation_da}, where the source corpus is 20News, and the backbone is NVDM with $K=50$. Based on these results, ``Highest to Similar'' obtains the highest quality in almost all the target domains, which shows the best generalisation capability among the DAs investigated. Thus, we apply ``Highest to Similar'' in Greg throughout our experiments.

\begin{table}[!t]
  \caption{Effect of Different Distances on Greg}
  \label{Ablation_dis}
  \centering
  \resizebox{0.5\textwidth}{!}{
  \begin{tabular}{cccccc}
    \toprule
    \multirow{3}{*}{(\%,$\uparrow$)} & \multirow{3}{*}{Method}  
     & \multicolumn{4}{c}{Targets} \\
     \cmidrule(l){3-6} 
    & &  Webs & TMN & DBpedia & R8 \\
    \midrule
    \multirow{4}{*}{CA} 
    & Euclidean Distance          & 41.1±1.1 & 39.7±0.2 & 38.0±0.9 & 69.3±0.6\\
    & Cosine Distance             & 24.1±0.3 & 26.9±0.2 & 14.6±0.3 & 59.5±0.6 \\
    & Hellinger Distance          & 38.8±1.0 & 38.8±0.6 & 31.3±0.6 & 73.3±0.9 \\
    & TopicalOT                        & \underline{63.7±0.3} & \underline{60.2±0.4} & \underline{55.8±0.4} & \underline{80.6±0.5}\\
    \midrule
    \multirow{4}{*}{TP} 
    & Euclidean Distance          & 30.0±0.4 & 32.1±0.6 & 21.0±0.5 & 61.5±1.3\\
    & Cosine Distance             & 23.4±0.1 & 25.8±0.1 & 11.6±0.4 & 55.3±0.6 \\
    & Hellinger Distance          & 28.8±0.4 & 32.3±0.5 & 19.6±0.2 & 62.0±2.0 \\
    & TopicalOT                        & \underline{35.8±1.5} & \underline{40.1±1.5} & \underline{26.3±0.9} & \underline{63.7±1.8}\\
    \midrule
    \multirow{4}{*}{TN} 
    & Euclidean Distance         & 6.5±0.2 & 3.8±0.2 & 9.9±0.5 & 12.4±1.1\\
    & Cosine Distance             & 4.3±0.1 & 2.2±0.2 & 5.4±0.3 & 10.0±0.4 \\
    & Hellinger Distance          & 6.5±0.3 & 3.7±0.2 & 9.0±0.4 & 12.9±1.1 \\
    & TopicalOT                        & \underline{12.7±0.8} & \underline{9.6±0.8} & \underline{15.7±0.5} & \underline{15.5±0.7}\\
  \bottomrule
\end{tabular}
}
\end{table}

\subsubsection{Effect of Distance Metrics on Greg}
Here, we demonstrate the effectiveness of TopicalOT in Greg in enhancing NTMs' generalisation power by changing different distance metrics. We consider other standard distances for the experiments, including Euclidean, Cosine and Hellinger distances. Their resulting topical representation quality of target corpora is illustrated in Table \ref{Ablation_dis}, where the source corpus is 20News, and the backbone is NVDM with $K=50$. Based on these results,  TopicalOT brings a significant improvement to the performance of the target domains and leaves a large margin compared to other distance metrics. It demonstrates the effectiveness of TopicalOT within our generalisation regularisation framework.

\begin{figure}[!t]
    \centering
\includegraphics[width=0.5\textwidth,height=\textheight,keepaspectratio]{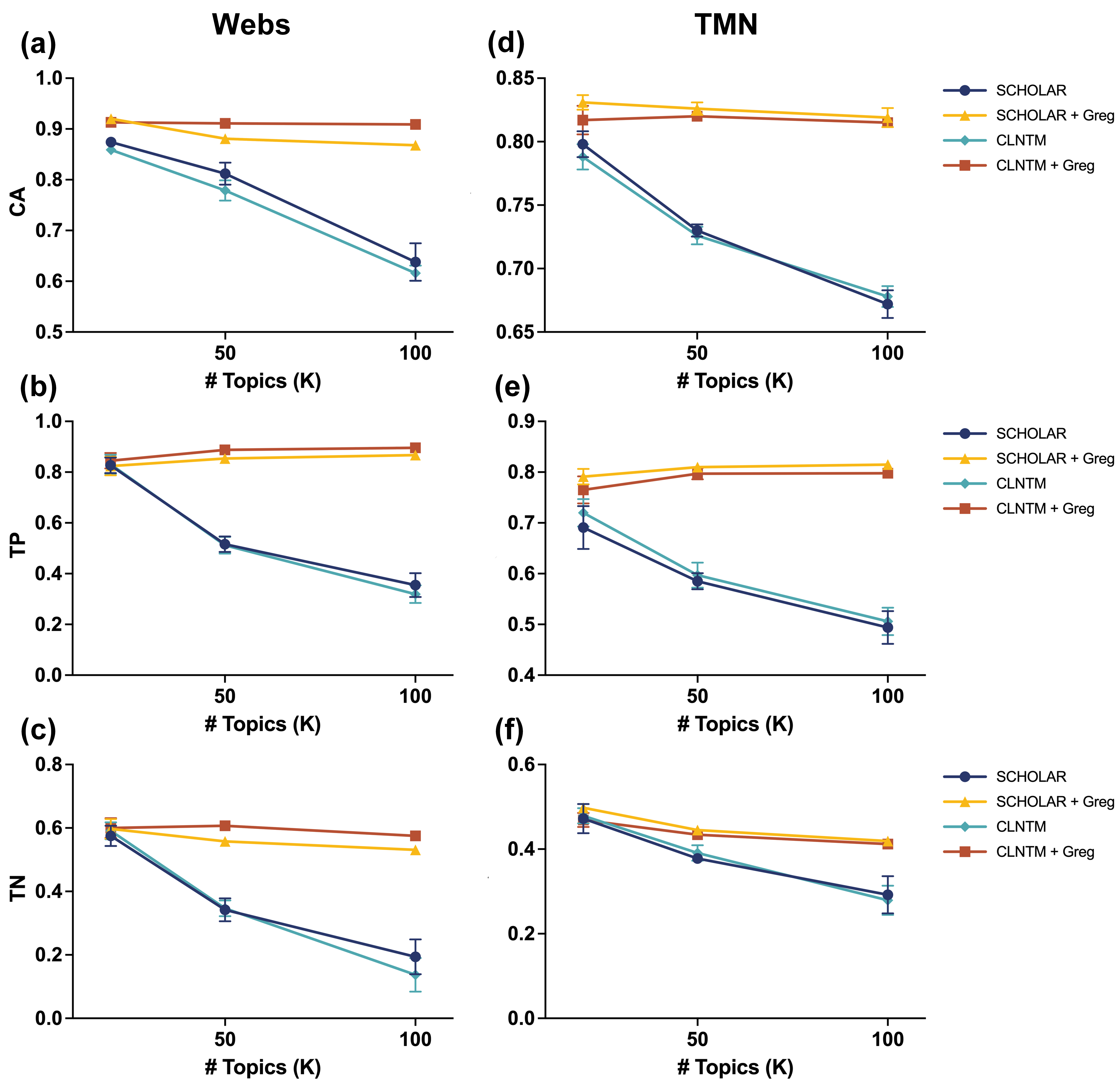}
    \caption{Effect of number of topics (i.e. $K$) to backbones and Greg. The quality of topical representation in terms of different metrics with different numbers of topics is illustrated in the figures.}
    \label{fig2}
\end{figure}

\subsubsection{Sensitivity to K in Short Document}
Setting the number of topics (i.e., $K$) for topic models is one of the challenges in short-text topic modelling \citep{xuan2016bayesian,qiang2020short}. From the experiments in previous sections, it can be noticed that Greg brings a huge improvement of topical representation quality for short documents when applying to SCHOLAR and CLNTM for most settings. Here, we explore the sensitivity to $K$ for backbone SCHOLAR and CLNTM with Greg. We plot the results for two short document corpora, Webs and TMN, in Figure \ref{fig2}. It can be observed that the topical representation quality of both SCHOLAR and CLNTM drops rapidly as $K$ increases, which indicates that they are sensitive to $K$ for short documents. Interestingly, their sensitivity issue is well addressed with Greg.

\subsubsection{Hyperparameter Sensitivity of Greg}
\begin{figure}[!t]
    \centering
\includegraphics[width=0.5\textwidth,height=\textheight,keepaspectratio]{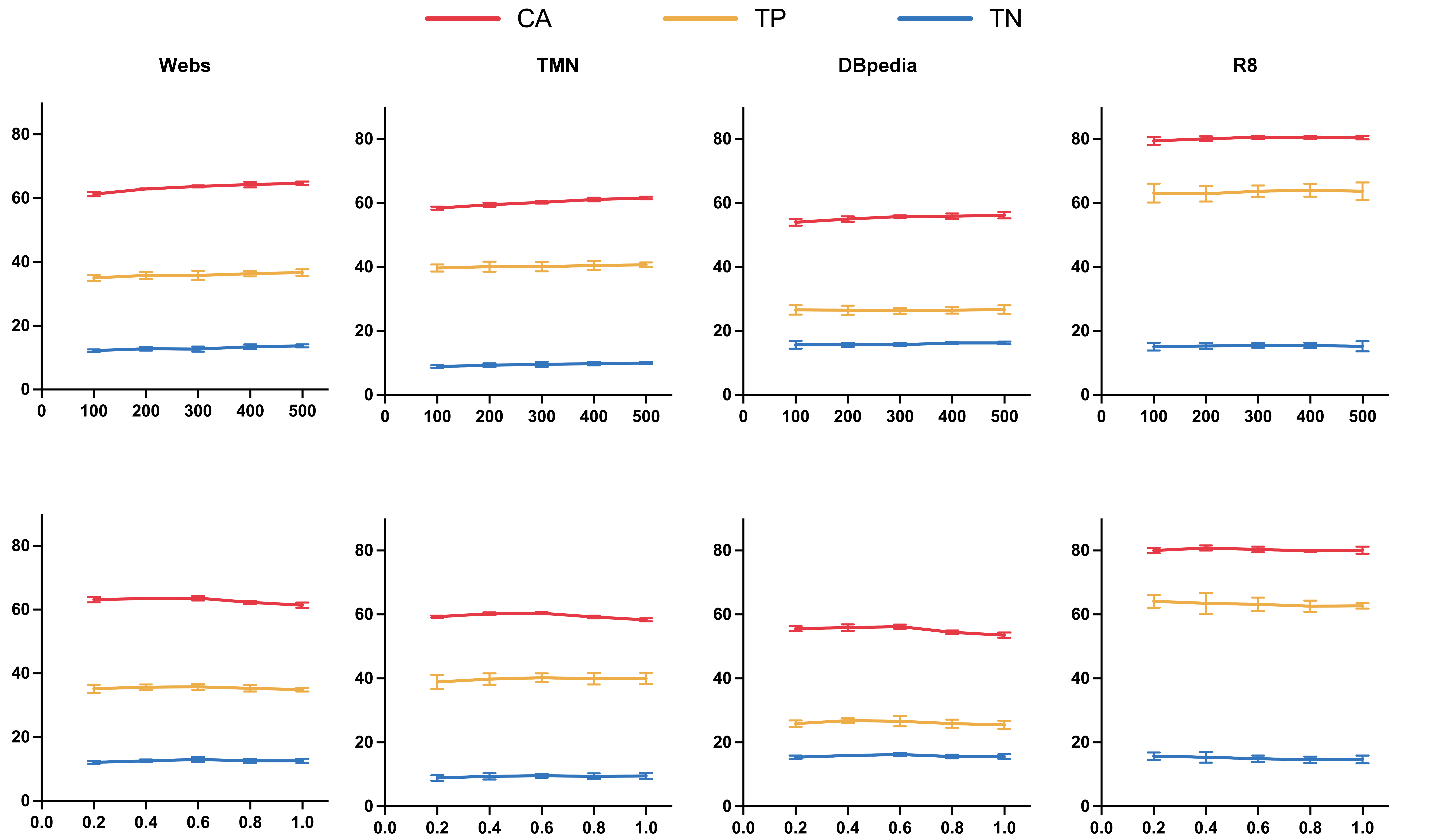}
    \caption{Hyperparameter Sensitivity of Greg. The x-axis of the first row is the regularisation weight $\gamma$; The x-axis of the second row is the augmentation rate $\beta$}
    \label{fig:hyper}
\end{figure}

Here, we study the sensitivity to the setting of hyperparameters in Greg, focusing on the regularisation weight $\gamma$ and augmentation weight $\beta$. We attach Greg to NVDM at $K=50$ on the `One Source to Different Targets' tasks as in Section \ref{sec:one_targets}. We vary the regularisation weight and augmentation rate, respectively, and record the performance of different metrics on different target corpora. Again, experiments are conducted 5 times with different random seeds. As shown in Figure \ref{fig:hyper}, whether varying the regularisation weight or the augmentation rate within a wide range, the benefits of Greg to target corpora performance still hold, and with little influence. It demonstrates Greg is not sensitive to the setting of its hyperparameters, thus its benefits are general, robust and reliable.

\begin{table}[!t]
  \caption{Example Topics}
  \label{topic_example}
  \resizebox{0.5\textwidth}{!}{
  \begin{tabular}{cc}
    \toprule
    Topic & Top 10 Words (first rows: PLDA, second rows: PLDA + Greg)\\
    \midrule 
    \multirow{2}{*}{ecosystems} 
    & habitat endemic tropical natural subtropical forest threatened loss moist ecuador\\
    & habitat endemic natural tropical forest subtropical loss moist threatened ecuador\\
    \cmidrule(l){2-2}
    \multirow{2}{*}{snail} 
    & snail marine gastropod sea mollusk family specie slug land terrestrial \\
    & snail sea gastropod marine mollusk family specie genus land terrestrial\\
    \cmidrule(l){2-2}
    \multirow{2}{*}{location} 
    & village mi east county km voivodeship central lie gmina poland \\
    & village county district population central mi gmina administrative voivodeship poland\\
    \cmidrule(l){2-2}
    \multirow{2}{*}{insects} 
    & moth wingspan family larva feed mm noctuidae tortricidae geometridae arctiidae \\
    & moth described arctiidae geometridae noctuidae family snout subfamily beetle genus\\
    \cmidrule(l){2-2}
    \multirow{2}{*}{religious} 
    & st church england century mary catholic saint parish paul roman \\
    & church st historic catholic parish street saint england mary place \\
    \cmidrule(l){2-2}
    \multirow{2}{*}{music}
    & album released music singer song band record musician songwriter rock\\
    & album released music record studio song singer band single debut \\
    \cmidrule(l){2-2}
    \multirow{2}{*}{river}
    & river long flow km mile tributary lake near creek source \\
    & river tributary long near mile flow km basin source creek\\
  \bottomrule
\end{tabular}
}
\end{table}

\subsection{Qualitative Analysis of Topics}
Although our primary focus is the generalisation of document representation, we show examples of the learned topics to understand what topics are captured after integrating Greg. We choose the learned topics on DBpedia using the backbone PLDA since it has the highest topic coherence (illustrated in Table S11) among all settings, making it easier to annotate topic labels. We pick the top coherent topics learned by PLDA and find their alignments in the learned topics of PLDA with Greg to explore the difference. The results are shown in Table \ref{topic_example}. From the ``location'' topic, we observe that the top words are more coherent and related to the administrative division after using Greg; From the ``music'' topic, Greg can discover words such as ``studio'', which may be less coherent to other music-related words.

\section{Conclusion}
In this work, we propose a new regularisation loss that can be integrated into many existing neural topic models (NTMs) for training on one dataset and generalising their topical representations to unseen documents without retraining. Our proposed loss, Greg, encourages NTMs to produce similar latent distributions for similar documents. The distance between document representations is measured by TopicalOT, which incorporates semantic information from both topics and words. Extensive experiments demonstrate that our framework, as a model-agnostic plugin for existing NTMs, significantly improves the generalisation ability of NTMs. In the future, we believe that topic model generalisation can be extended to generalising both document representations and topics across different languages and modalities.

\bibliographystyle{ieeetr}
\bibliography{main.bib}

\end{document}